\documentclass[runningheads]{llncs}

\usepackage{eccv}
\usepackage{eccvabbrv}

\usepackage{graphicx}
\usepackage{wrapfig}
\usepackage{booktabs}
\usepackage[table]{xcolor}
\usepackage{array}
\usepackage[table]{xcolor}
\usepackage{collcell}
\usepackage{hhline}
\usepackage{caption}

\usepackage{multirow}
\usepackage[T1]{fontenc} 
\usepackage{underscore}  
\usepackage{longtable}
\usepackage{colortbl}
\usepackage{tcolorbox}
\tcbuselibrary{listings, breakable, skins}
\tcbset{
  promptbox/.style={
    enhanced,
    breakable,
    colback=gray!8,
    colframe=gray!30,
    colbacktitle=black,
    coltitle=white,
    fonttitle=\bfseries\small,
    fontupper=\small,
    left=8pt, right=8pt, top=4pt, bottom=6pt,
    boxrule=0.4pt,
    arc=3pt,
    toptitle=1pt,
    bottomtitle=1pt,
  }
}

\newcolumntype{H}{>{\collectcell\applycolor}c<{\endcollectcell}}
\usepackage[accsupp]{axessibility}  % Improves PDF readability for those with disabilities.

\usepackage{hyperref}

\begin{document}

% ---------------------------------------------------------------
\title{TriViewBench: Controlled Complexity Scaling \texorpdfstring{\\}{ }for Multi-View Structural Reasoning in MLLMs}

\titlerunning{TriViewBench}

\author{Yu-Yang Chen\inst{1,2} \and
Lan-Zhe Guo\inst{1,2}}

\authorrunning{Y.-Y. Chen and L.-Z. Guo}

\institute{School of Intelligence Science and Technology, Nanjing University, China \and
National Key Laboratory for Novel Software Technology, Nanjing University, China
\email{yuyangchen@smail.nju.edu.cn, guolz@lamda.nju.edu.cn}}
\maketitle

\begin{abstract}
  Multimodal Large Language Models (MLLMs) demonstrate strong performance on standard visual question answering benchmarks, yet their scalability under controlled structural complexity remains poorly understood. We introduce \textbf{TriViewBench}, a controlled three-view visual reasoning benchmark constructed from synthetic 3D scenes with explicitly parameterized object count and occlusion. The benchmark contains 1,923 scenes and over 14K Question-Answer (QA) pairs organized into four complexity levels and three reasoning categories: \textbf{Local Decision}, \textbf{Object Counting}, and \textbf{Global Recovery}. We evaluate 18 open- and closed-source MLLMs under a unified prompting protocol. All 18 models exhibit an identical capability hierarchy without exception (Local Decision $>$ Object Counting $>$ Global Recovery), and performance degrades monotonically with complexity: Local Decision tasks decline modestly (12.11\% relative drop), while Object Counting degrades substantially (59.14\%) and Global Recovery collapses severely (80.02\%). Error analysis on Object Counting reveals two mechanistically independent failure modes: single-view tasks are dominated by undercounting due to occlusion blindness, whereas the multi-view task reverses to overcounting due to cross-view identity confusion. Chain-of-Thought (CoT) prompting yields near-zero overall benefit ($\Delta = -0.16\%$) and its effect on Global Recovery is strongly capability-gated, suggesting that the bottleneck lies in cross-view spatial representation rather than reasoning strategy. These findings reveal fundamental scalability limitations in current MLLMs and position TriViewBench as a controlled diagnostic framework for analyzing structural reasoning failures.
\end{abstract}

\section{Introduction}

\begin{figure}[t]
    \centering
    \includegraphics[width=\linewidth, trim=10 50 70 10, clip]{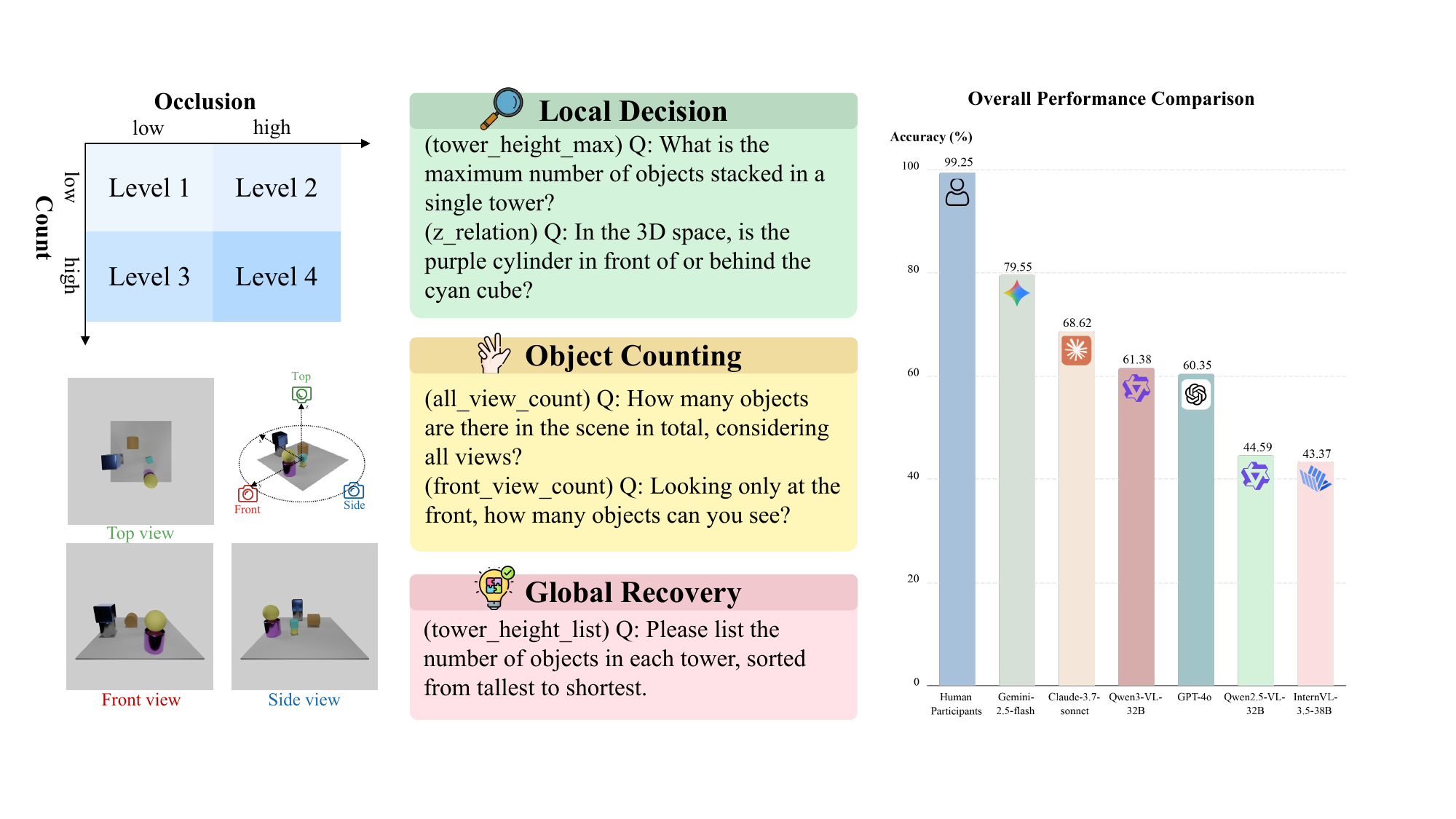} 
    \caption{
    Overview of TriViewBench. 
    \textbf{Left}: Four complexity levels defined by object count and occlusion density with three-view rendering. 
    \textbf{Center}: Examples of questions in three reasoning categories (Local Decision, Object Counting, and Global Recovery). 
    \textbf{Right}: Overall performance comparison between humans and representative MLLMs.
    }
    \label{fig:overview}
\end{figure}

Multimodal Large Language Models (MLLMs) achieve strong performance across various visual question answering tasks~\cite{yin2024survey, kuang2025natural, jin2025efficient}.
In multi-view settings, however, reliable reasoning demands more than recognizing individual objects: models must align identities across views, resolve occlusion by using complementary viewpoints,
and reconstruct coherent structural representations from fragmented visual cues~\cite{yeh2025seeing, yang2025mmsi, park2025nuplanqa}.
Whether current MLLMs can maintain this level of reasoning as scene complexity increases in a controlled and measurable way remains an open question.

Existing visual reasoning benchmarks provide valuable progress but exhibit several structural limitations that make this question difficult to answer~\cite{wang2025spatial457, ma20253dsrbench, lee2025spatialmosaic}.
Most benchmarks are constructed from unconstrained real-world images~\cite{daxberger2025mm, park2025nuplanqa, yang2025mmsi, lee2025spatialmosaic}.
While such data offers diversity, scene complexity is not explicitly parameterized: object count, spatial arrangement, and occlusion degree vary simultaneously and unpredictably, so performance degradation cannot be cleanly attributed to specific factors.
Most benchmarks also operate in single-view settings, which limit the demand for cross-perspective integration~\cite{johnson2017clevr, ma20253dsrbench, gong2025space10}.
Furthermore, real-image benchmarks typically lack precise object-level geometry and visibility annotations, preventing fine-grained decomposition of errors into local, counting, or structural failures.

To address these limitations, we introduce \textbf{TriViewBench}, a controlled multi-view benchmark built from synthetic 3D scenes with explicitly parameterized object count and occlusion.
Each scene is rendered from front, side, and top-down viewpoints,
enforcing cross-view reasoning and structural consistency.
The benchmark contains 1,923 scenes and over 14K Question-Answer (QA) pairs organized into four complexity levels and three reasoning categories:
\textbf{Local Decision}, \textbf{Object Counting}, and
\textbf{Global Recovery}.
By varying complexity axes independently and structuring tasks by reasoning demand, TriViewBench enables direct measurement of performance scaling and interpretable failure analysis.

\Cref{fig:overview} provides an overview of the benchmark design,
question taxonomy, and performance comparison between humans and representative MLLMs.

We evaluate 18 open- and closed-source MLLMs under a unified prompting protocol.
The results reveal four consistent findings.
First, all models exhibit an identical capability hierarchy: Local Decision $>$ Object Counting $>$ Global Recovery,
forming a stable gradient across architectures,
parameter scales, and model families.
Second, performance degrades monotonically with complexity at markedly different rates across categories: Local Decision tasks decline modestly (12.11\% relative drop from Level 1 to 4), Object Counting degrades substantially (59.14\%), and Global Recovery deteriorates severely (80.02\% relative drop).
Third, error analysis on Object Counting uncovers two mechanistically distinct failure modes:
single-view tasks are dominated by undercounting, as models fail to accurately perceive partially occluded objects from a single view;
the multi-view task reverses to overcounting, as models lose cross-view object identity and count the same physical object multiple times.
These two modes are comparable in magnitude but opposite in direction,
confirming that they arise from independent perceptual mechanisms.
Fourth, Chain-of-Thought (CoT) prompting yields negligible overall gain ($\Delta = -0.16\%$), with its effect on Global Recovery being strongly capability-gated: large open-source models ($\geq$14B) improve by an average of $+17.03\%$, while small models ($\leq$3B) uniformly deteriorate.
Even with CoT, no model approaches human performance,
indicating that the root limitation is cross-view spatial representation rather than reasoning~strategy.

In summary, the main contributions of this work are:
\begin{itemize}
  \item We introduce TriViewBench, a controlled multi-view benchmark with
    independently parameterized object count and occlusion, providing a
    systematic diagnostic framework for structural reasoning under
    complexity scaling.
  \item We establish a universal capability hierarchy across all 18
    evaluated MLLMs (Local Decision $>$ Object Counting $>$ Global
    Recovery) and quantify the differential degradation rates across
    categories.
  \item We identify two mechanistically independent counting failure modes, namely single-view occlusion blindness and cross-view identity confusion, through error direction analysis. These findings reveal distinct perceptual bottlenecks that emerge at comparable magnitudes but in opposite directions.
  \item We show that CoT prompting does not resolve structural scaling
    failures and that its benefit on Global Recovery is capability-gated,
    pointing to cross-view spatial representation as the fundamental
    bottleneck.
\end{itemize}

% ----------------------------------------------------------------
\section{Related Work}
% ----------------------------------------------------------------

\paragraph{Multimodal Large Language Models.}
Recent MLLMs combine vision encoders with large language models to perform diverse visual tasks,
including image captioning, visual question answering, and scene understanding~\cite{comanici2025gemini, hurst2024gpt, liu2023visual, liu2024improved, li2024llava, bai2025qwen3, wang2025internvl3}.
Model scale and architectural design have been shown to influence reasoning capability, though their relative contributions remain contested~\cite{bai2025qwen3,wang2025internvl3,bai2025qwen25vltechnicalreport}.
TriViewBench tests these factors by evaluating 18 models spanning a wide range of scales and architectures under identical conditions, finding that architectural family predicts performance more reliably than parameter count alone.

\paragraph{Visual Reasoning Benchmarks.}
Most existing visual reasoning benchmarks are built around single images~\cite{ma20253dsrbench, antol2015vqa, marino2019ok, yue2024mmmu, li2023seed},
which limits the demand for cross-view information integration:
models need only process one perspective at a time and are never required to reconcile observations across viewpoints or resolve occlusion by consulting complementary angles.
A growing body of work has extended evaluation to multi-image or multi-view settings to address this gap.
MuirBench~\cite{wang2024muirbench} covers 12 multi-image task categories and finds that even GPT-4o achieves only 68.0\% accuracy.
MMSI-Bench~\cite{yang2025mmsi} targets multi-image spatial intelligence and finds that the strongest open-source model attains only 30.7\% accuracy against a human baseline of 97.2\%, with six domain experts spending over 300 hours crafting
1,000 questions manually.
All-Angles Bench~\cite{yeh2025seeing} evaluates multi-view understanding across real-world scenes and identifies cross-view occlusion as a central failure mode.
These benchmarks demonstrate that multi-view reasoning poses genuine challenges to current MLLMs, but their reliance on real-world imagery introduces two interrelated limitations.
First, collecting and annotating multi-view real-world data is labor-intensive: establishing ground-truth answers for spatial relations, occlusion states, and structural attributes requires careful manual verification for each scene.
Second, and more fundamentally, scene complexity in real images is uncontrolled. Object count, spatial arrangement, and occlusion degree vary simultaneously and without explicit parameterization.
When accuracy drops, it is impossible to attribute failure to specific causal factors or to measure how performance scales with individual complexity dimensions.
TriViewBench addresses both limitations through synthetic,
parameterized scene generation: ground-truth annotations are derived automatically from scene metadata without any manual labeling, and complexity axes are varied independently,
enabling direct attribution of observed failures to object cardinality, occlusion severity, or their combination.

\paragraph{Synthetic and Controllable Benchmarks.}
CLEVR~\cite{johnson2017clevr} showed that controlled synthetic generation enables precise failure diagnosis invisible to real-image evaluations, establishing a paradigm that has since informed numerous diagnostic benchmarks.
In the domain of Object Counting, VLMCountBench~\cite{guo2025your}
shows that MLLMs fail under compositional counting scenarios, consistent with our findings.
TriViewBench extends this controlled paradigm to a three-view setting, adding occlusion as an explicit complexity axis and introducing a cross-view structural reasoning dimension absent from prior synthetic benchmarks.

\paragraph{Spatial Reasoning and CoT Prompting.}
Prior work probes spatial reasoning from complementary angles:
3DSRBench~\cite{ma20253dsrbench} exposes degradation under non-canonical viewpoints across 12 question types,
Spatial457~\cite{wang2025spatial457} diagnoses 6D spatial reasoning through a cascading evaluation structure across five difficulty levels,
and MM-Spatial~\cite{daxberger2025mm} advances 3D spatial understanding via multi-view and metric depth inputs, targeting spatial relationship prediction, metric estimation, and 3D grounding.
These benchmarks reveal important limitations in model spatial awareness,
but focus primarily on metric estimation, orientation, and 3D grounding rather than the structural reconstruction and cross-view identity resolution that characterize TriViewBench's task design.
CoT prompting~\cite{wei2022chain,zhang2023multimodal,mitra2024compositional} has been shown to improve multi-step reasoning in language tasks.
Our evaluation finds that CoT yields near-zero overall benefit ($\Delta = -0.16\%$) and that its effect on the hardest category is strongly capability-gated, pointing to spatial representation rather than reasoning strategy as the primary bottleneck.

\section{TriViewBench}
\begin{figure}[t]
    \centering
    \includegraphics[width=\linewidth]{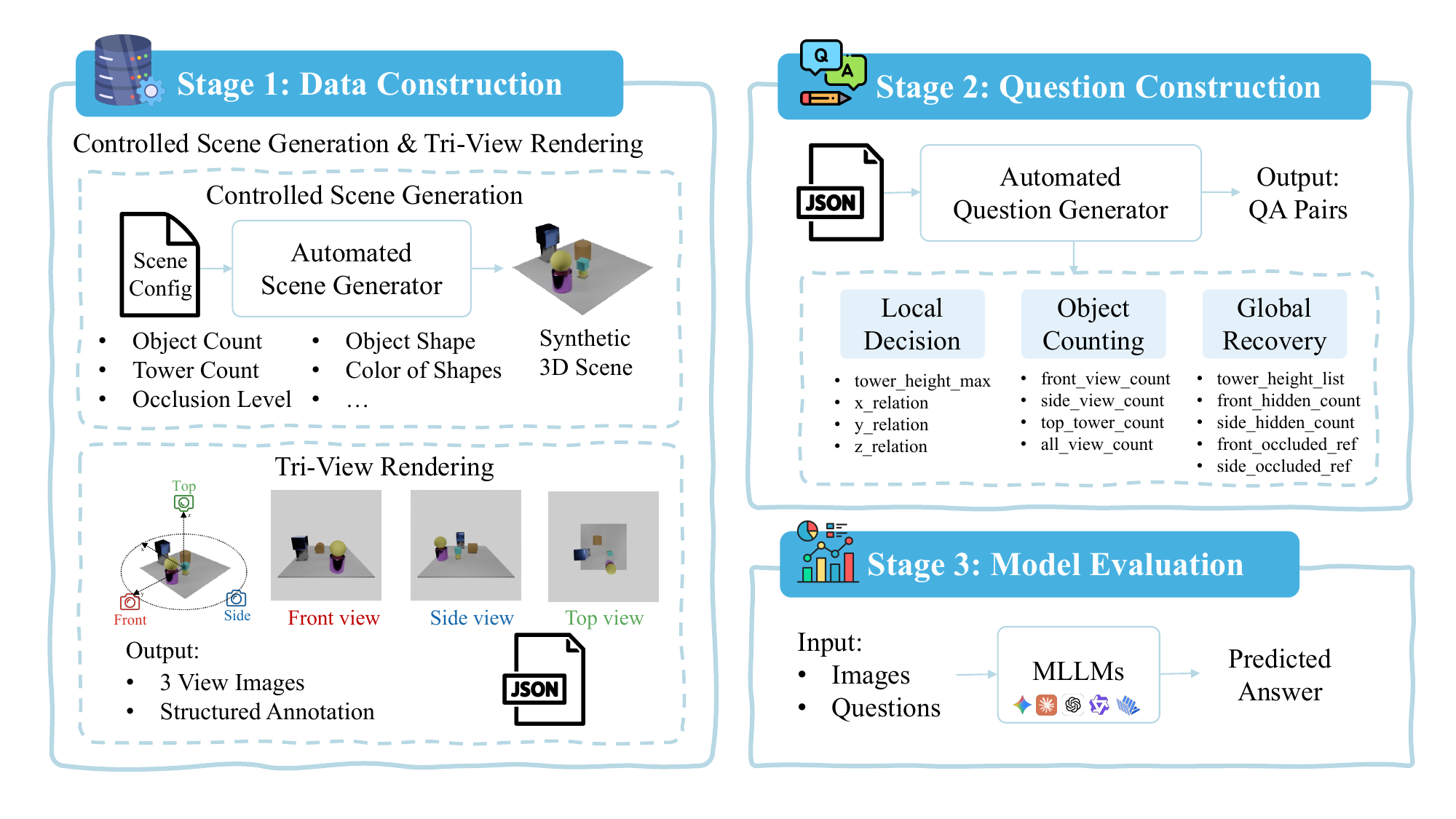} 
    \caption{Illustration of the TriViewBench construction pipeline. The workflow comprises three main stages: (1) \textbf{Data Construction}: generating synthetic 3D scenes from parameterized configs and rendering three-view images with structured annotations; (2) \textbf{Question Construction}: automatically synthesizing QA pairs across three reasoning Categories (Local Decision, Object Counting, and Global Recovery) based on scene metadata; (3) \textbf{Model Evaluation}: benchmarking 18 MLLMs under various complexity levels and prompting protocols.}
    \label{fig:pipeline}
\end{figure}

In this section, we present TriViewBench, a controlled benchmark for evaluating multi-view structural reasoning. We first introduce the benchmark construction process in \cref{subsec:construction}, and then provide an overview of its task organization and distribution in \cref{subsec:overview}.

\subsection{Benchmark Construction Process}
\label{subsec:construction}

As illustrated in \cref{fig:pipeline}, TriViewBench is built through three stages: data construction, question construction, and model evaluation.

\subsubsection{Data Construction.}
Scenes are generated using a parameterized pipeline built on top of Kubric~\cite{greff2022kubric}, an open-source framework for procedural scene generation.
Each scene is controlled by two independent axes of complexity,
yielding four levels:
\textbf{Level 1} (LowCount\_LowOcclusion),
\textbf{Level 2} (LowCount\_HighOcclusion),
\textbf{Level 3} (HighCount\_LowOcclusion), and
\textbf{Level 4} (HighCount\_HighOcclusion).
The count axis distinguishes \textit{LowCount} scenes (1--5 objects) from \textit{HighCount} scenes (6--10 objects).
The occlusion axis is defined through the front and side views only,
since vertical stacking can produce near-zero top-down visibility,
and the top view is conventionally reserved for spatial localization.
\textit{LowOcclusion} requires all objects to have visibility ratios above 60\% in both lateral views.
\textit{HighOcclusion} requires at least one object to fall below 30\% visibility in either view; crucially, the same object must exceed 60\% visibility in the other view, ensuring that every question remains answerable from the provided images.
For each scene, three virtual cameras render aligned front, side, and top-down images.
A structured JSON annotation is generated alongside,
recording a rich set of metadata including world coordinates,
camera parameters, object properties,
spatial relations between objects,
and per-view bounding boxes and visibility ratios.
These annotations serve as the sole ground truth for all downstream QA generation, guaranteeing answer correctness and uniqueness by construction.

\subsubsection{Question Construction.}
QA pairs are generated automatically from the structured annotations using deterministic templates, with no manual labeling required.
We define 13 fine-grained task types grouped into three categories.
\textbf{Local Decision} covers pairwise spatial relation judgment (relative position along x/y/z axes) and comparative attribute queries (identifying the tallest tower), both of which can be resolved from locally visible objects without requiring complete scene enumeration.
\textbf{Object Counting} requires fine-grained enumeration within or across views, including single-view counting, tower-level counting from the top perspective, and cross-view identity matching.
\textbf{Global Recovery} demands comprehensive cross-view synthesis,
with the primary task being full tower height distribution recovery;
this category also includes occlusion-specific sub-tasks (hidden object counting and occluded object identification) that appear exclusively in HighOcclusion levels (Level 2 and Level 4).
Because all answers are derived deterministically from annotations,
ambiguity and labeling error are eliminated by design.

\subsubsection{Model Evaluation.}
In the evaluation stage, the three rendered images and a question are jointly provided to the model.
The textual response is compared against the ground-truth answer derived from the annotation.

\subsection{Overview of TriViewBench}
\label{subsec:overview}

\begin{table}[t] 
    \centering
    \begin{minipage}[t]{0.68\linewidth} 
        \centering
        \footnotesize
        \caption{Taxonomy of reasoning tasks. $\dagger$ denotes sub-types that appear only in
          HighOcclusion levels (Level 2 and 4).}
        \label{tab:taxonomy}
        \setlength{\tabcolsep}{4pt}
        \resizebox{\linewidth}{!}{
            \begin{tabular}{lll}
                \toprule
                \textbf{Category} & \textbf{Sub-type} & \textbf{Description} \\
                \midrule
                \textbf{Local} & tower\_height\_max & Identify the maximum tower height. \\
                \textbf{Decision} & x/y/z\_relation & Relative 3D spatial relations. \\
                \midrule
                \textbf{Object} & front/side\_view\_count & Single-view visible object counting. \\
                \textbf{Counting} & top\_tower\_count & Count towers from top perspective. \\
                & all\_view\_count & Cross-view identity matching. \\
                \midrule
                \textbf{Global} & tower\_height\_list & Full height distribution recovery. \\
                \textbf{Recovery} & front/side\_hidden\_count$^\dagger$ & Reason about fully occluded objects. \\
                & front/side\_occluded\_ref$^\dagger$ & Identify partially obscured objects. \\
                \bottomrule
            \end{tabular}
        }
    \end{minipage}
    \hfill
    \begin{minipage}[t]{0.3\linewidth} 
        \centering
        \makeatletter\def\@captype{figure}\makeatother 
        \raisebox{-110pt}{\includegraphics[width=\linewidth, trim=140 125 140 140, clip]{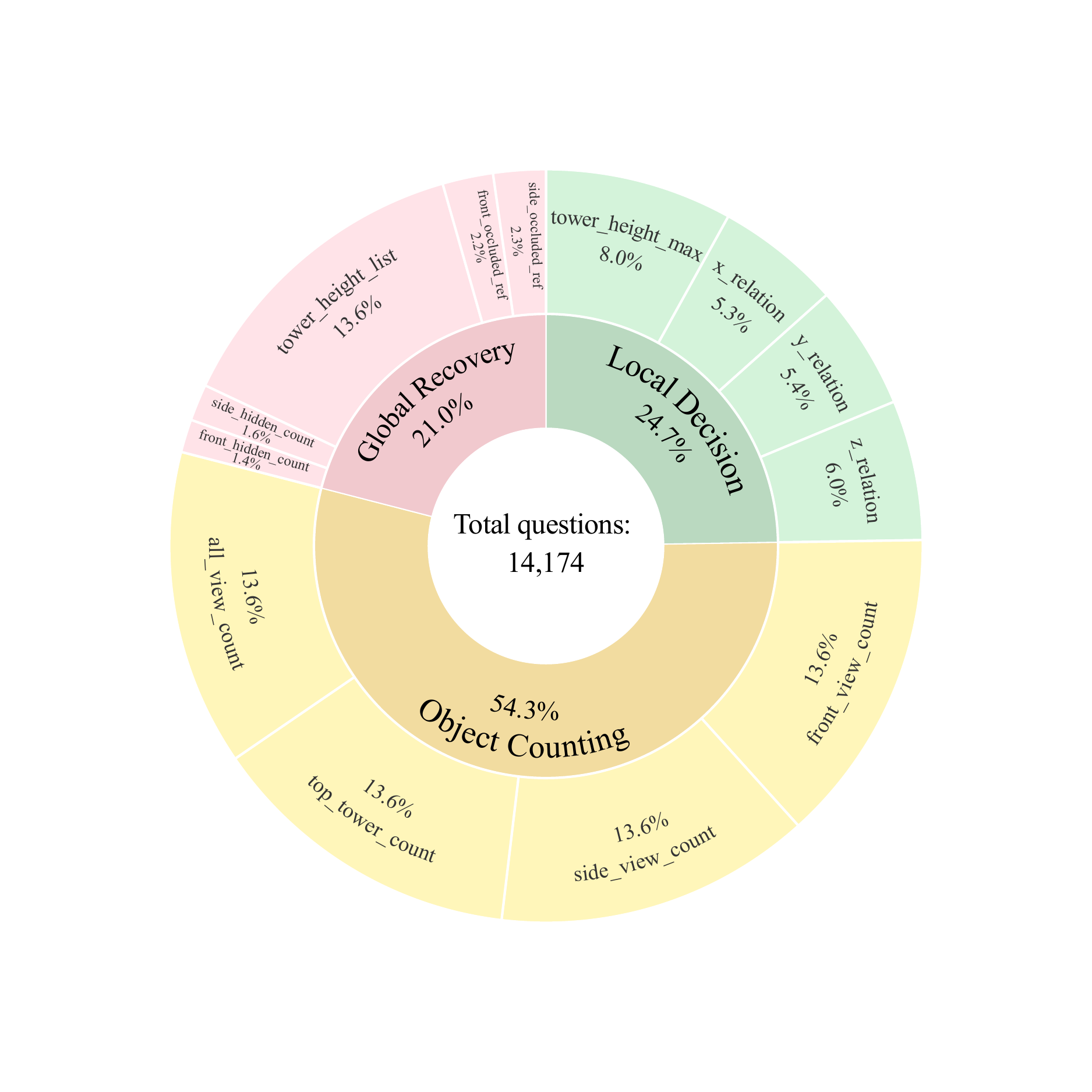}}
        \caption{Task distribution.}
        \label{fig:percent}
    \end{minipage}
\end{table}

TriViewBench contains 14,174 QA pairs spanning 13 reasoning sub-types organized into three categories.
The task taxonomy is summarized in \cref{tab:taxonomy},
and the distribution across categories is shown in \cref{fig:percent}.

Object Counting forms the largest portion (slightly over half),
distributed evenly across single-view counting, cross-view identity matching, and top-view tower counting, ensuring balanced coverage of different aggregation scenarios.
Local Decision constitutes approximately one quarter of the dataset,
covering spatial relations along three axes and maximum tower height identification.
Global Recovery accounts for the remainder, with full tower height
reconstruction as its primary sub-task; occlusion-specific sub-types
appear only in Level 2 and Level 4, forming a smaller subset.
The two complexity axes, object count and occlusion severity,
are varied independently, enabling controlled attribution of
performance changes to each factor separately or their combination.

\begin{table}[t]
\centering
\caption{Evaluation results for 18 MLLMs on TriViewBench. For each level, the best-performing proprietary model and the best-performing open-source model are both indicated in \textbf{bold}.}
\label{tab:main_results}
% \begin{tabular}{l cccc c}
\begin{tabular*}{0.9\linewidth}{@{\extracolsep{\fill}} l wc{1.1cm} wc{1.1cm} wc{1.1cm} wc{1.1cm} wc{1.3cm} @{}}
\toprule
 & \multicolumn{4}{c}{Complexity Levels} & \\
\cmidrule(lr){2-5}
Model & Level 1 & Level 2 & Level 3 & Level 4 & \textbf{Overall} \\
\midrule
\rowcolor[gray]{.95} \multicolumn{6}{l}{\textit{Proprietary}} \\
GPT-4o                           & 86.41       & 68.70       & 50.74       & 42.76       & 60.35            \\
Claude-3.7-Sonnet                & 84.16       & 72.41       & 67.99       & 54.90       & 68.62            \\
Gemini-2.5-Flash                 & \textbf{91.52}       & \textbf{83.35}       & \textbf{81.90}       & \textbf{65.94}       & \textbf{79.55}            \\
\rowcolor[gray]{.95} \multicolumn{6}{l}{\textit{Open-source}} \\
LLaVA-OneVision-Qwen2-0.5B       & 18.15       & 21.64       & 14.69       & 15.91       & 17.48            \\
LLaVA-OneVision-Qwen2-7B         & 45.60       & 28.17       & 18.07       & 18.34       & 26.49            \\
Qwen2.5-VL-3B-Instruct           & 47.21       & 28.40       & 14.34       & 20.37       & 26.59            \\
Qwen2.5-VL-7B-Instruct           & 64.14       & 42.36       & 22.74       & 25.05       & 37.05            \\
Qwen2.5-VL-32B-Instruct          & 67.21       & 49.38       & 35.30       & 32.10       & 44.59            \\
Qwen3-VL-2B-Instruct             & 60.34       & 48.10       & 38.61       & 33.59       & 44.07            \\
Qwen3-VL-4B-Instruct             & 69.55       & 58.80       & 43.61       & 40.75       & 52.00            \\
Qwen3-VL-8B-Instruct             & 79.54       & 64.93       & \textbf{55.58}       & \textbf{46.68}       & 60.32            \\
Qwen3-VL-32B-Instruct            & \textbf{86.04}       & \textbf{69.39}       & 54.26       & 43.02       & \textbf{61.38}            \\
InternVL3.5-1B                  & 47.18       & 34.32       & 24.94       & 25.90       & 32.26            \\
InternVL3.5-2B                  & 56.78       & 38.53       & 29.55       & 26.87       & 36.76            \\
InternVL3.5-4B                  & 49.39       & 45.66       & 40.35       & 34.11       & 41.71            \\
InternVL3.5-8B                  & 40.84       & 40.91       & 37.52       & 28.43       & 36.34            \\
InternVL3.5-14B                 & 51.86       & 46.68       & 39.54       & 32.45       & 41.79            \\
InternVL3.5-38B                 & 56.55       & 46.19       & 39.28       & 34.98       & 43.37            \\
\rowcolor[gray]{.95} \multicolumn{6}{l}{\textit{Human}} \\
Human Performance & 99.67 & 99.00 & 99.67 & 98.67 & 99.25 \\
\bottomrule
\end{tabular*}
\end{table}

\section{Experiments}

% ----------------------------------------------------------------
\subsection{Evaluation Setup}
% ----------------------------------------------------------------

\paragraph{Models.}
We evaluate 18 MLLMs spanning proprietary and open-source systems across a broad range of architectures and parameter scales.
Proprietary models include GPT-4o~\cite{hurst2024gpt},
Gemini-2.5-Flash~\cite{comanici2025gemini}, and Claude-3.7-Sonnet~\cite{anthropic2025claude37}.
Open-source models include two LLaVA-OneVision variants (0.5B, 7B)~\cite{li2024llava}, three Qwen2.5-VL variants (3B--32B)~\cite{bai2025qwen25vltechnicalreport}, four Qwen3-VL variants (2B--32B)~\cite{bai2025qwen3},
and six InternVL3.5 variants (1B--38B)~\cite{wang2025internvl3}.

\paragraph{Inference regimes.}
Two prompting protocols are evaluated.
\textbf{Direct} prompting instructs models to produce a concise final answer within a 128-token limit, measuring direct perceptual and structural reasoning without explicit intermediate reasoning.
\textbf{CoT} prompting~\cite{wei2022chain} allows extended intermediate reasoning up to 2,048 tokens.
All experiments use greedy decoding (temperature~$=0$) to ensure reproducibility.

\paragraph{Evaluation protocol.}
Accuracy is the primary metric across all tasks.
For most question types, correctness is determined by exact string match.
For Global Recovery (tower\_height\_list), which requires predicting an ordered list of tower heights, predictions and ground-truth lists are both sorted in descending order before comparison; a response is marked correct only if the full sorted sequence agrees exactly.

\paragraph{Human baseline.}
Three participants independently answered a stratified random subset of 400 questions (100 per complexity level), achieving an overall accuracy of 99.25\%.

\subsection{Main Results}

\Cref{tab:main_results} presents the full benchmark results. We summarize our main findings as follows:

\textbf{All models degrade substantially as structural complexity increases.}
Across all 18 evaluated models, accuracy declines monotonically from Level~1 to Level~4.
At Level~1, several models exceed 85\% accuracy;
by Level~4, most fall below the mid-40\% range.

\textbf{Human performance remains near ceiling, exposing a persistent gap.}
Human participants maintain high accuracy across all levels,
revealing that the structural challenges in our benchmark are well within human capacity but remain fundamentally difficult for current MLLMs.

\textbf{Proprietary scale does not eliminate structural scaling failures.}
Among proprietary systems, Gemini-2.5-Flash achieves the highest overall accuracy but still degrades substantially across levels (91.52\%~$\to$~65.94\%).
GPT-4o and Claude-3.7-Sonnet follow similar trends,
confirming that closed-source scale alone cannot resolve compositional reasoning deficits.

\textbf{Larger models help, but with diminishing and unstable gains.}
Among open-source models, Qwen3-VL models perform comparably to GPT-4o and Claude-3.7-Sonnet.
Within model families, larger parameter counts generally improve performance,
but the marginal gains diminish as models grow larger.
Moreover, the advantage of larger variants narrows as complexity increases,
and in some cases smaller models outperform larger counterparts at higher levels,
suggesting that scale alone does not confer compositional robustness.

\textbf{Architecture and training matter more than parameter count.}
Cross-family comparisons support this claim.
Qwen2.5-VL-3B-Instruct matches LLaVA-OneVision-Qwen2-7B despite having less than half the parameters,
and Qwen3-VL-2B-Instruct performs on par with Qwen2.5-VL-32B-Instruct and InternVL3.5-38B.
These observations indicate that training strategy and architecture contribute more to structural reasoning capability than model scale alone.

% ----------------------------------------------------------------
\subsection{Category-Level Analysis}
\label{sec:category}
% ----------------------------------------------------------------

\begin{table}[t]
\centering
\begin{minipage}[t]{0.48\linewidth}
  \centering
  \footnotesize
  \caption{
    \textbf{Category ranking distribution} across all 18 models.
    Category-level accuracy is averaged over four complexity levels.
    All models exhibit an identical hierarchy without exception.
  }
  \label{tab:rank}
  \resizebox{\linewidth}{!}{
  \renewcommand{\arraystretch}{1.3}
  \begin{tabular}{lccc}
    \toprule
    \textbf{Category} & \textbf{Rank 1} & \textbf{Rank 2} & \textbf{Rank 3} \\
    \midrule
    Local Decision   & 18 & 0  & 0  \\
    Object Counting  & 0  & 18 & 0  \\
    Global Recovery  & 0  & 0  & 18 \\
    \bottomrule
  \end{tabular}}
\end{minipage}
\hfill
\begin{minipage}[t]{0.48\linewidth}
  \centering
  \makeatletter\def\@captype{figure}\makeatother
  \raisebox{-100pt}{\includegraphics[width=\linewidth,
    trim=10 10 10 10, clip]{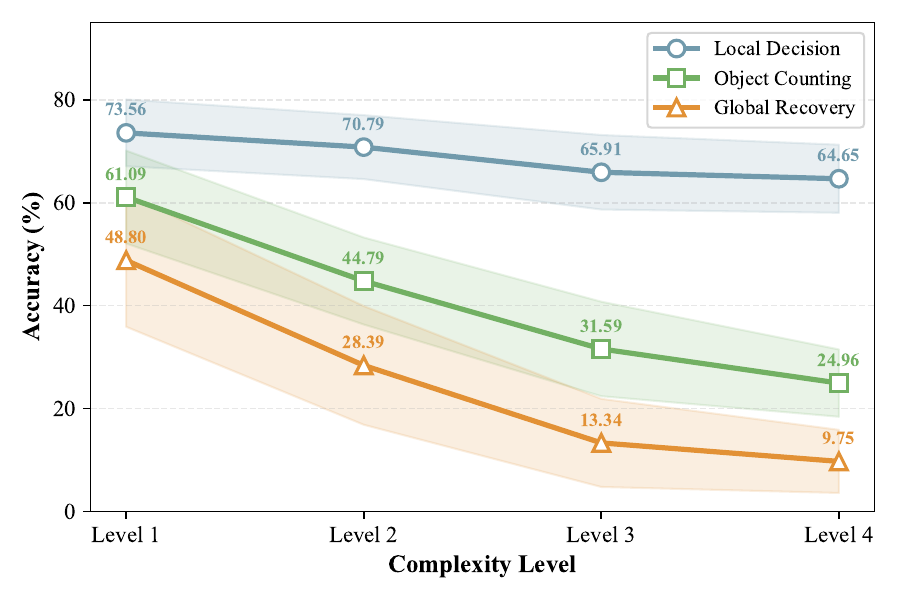}}
  \caption{
    \textbf{Category-level performance.} Lines: average accuracy over 18 MLLMs; Shaded areas: 95\% confidence intervals.
  }
  \label{fig:category}
\end{minipage}
\end{table}

To identify which reasoning demands drive performance degradation,
we analyze the three categories separately.
We restrict this analysis to question types present across all four complexity levels,
ensuring that observed changes reflect structural complexity scaling rather than task composition shifts.
The four question types exclusive to Levels~2 and~4 are analyzed separately in \cref{sec:occlusion_specific}.

\textbf{All 18 models share an identical category ranking without exception.} \Cref{tab:rank} shows the category ranking distribution across all evaluated models.
Every model ranks Local Decision first, Object Counting second, and Global Recovery third.
The unanimity of this ordering across architectures of varying scale and design establishes it as a systematic capability gradient rather than a model-specific artifact.
Given this consistency, we report model-averaged results below;
per-model breakdowns are provided in Appendix.

\textbf{The three categories degrade at fundamentally different rates.}
% \Cref{fig:category} quantifies the scaling behavior of each category.
\Cref{fig:category} quantifies each category's scaling behavior.
All three degrade monotonically with increasing complexity,
but the gap between them widens dramatically.
Local Decision is the most resilient, decreasing from 73.56\% to 64.65\% (8.91~pp absolute, 12.11\% relative).
Object Counting suffers moderate degradation, declining from 61.09\% to 24.96\% (36.13~pp absolute, 59.14\% relative).
Global Recovery is the most fragile, collapsing from 48.80\% to 9.75\% (39.05~pp absolute, 80.02\% relative),
indicating that holistic structural reconstruction is disproportionately vulnerable to increasing complexity.
% Limited confidence-interval overlap confirms that these differential degradation rates are robust rather than sampling artifacts.
Limited confidence-interval overlap confirms these differential degradation rates are robust rather than sampling artifacts.

% ----------------------------------------------------------------
\subsection{Analysis}
% ----------------------------------------------------------------

\subsubsection{Effect of Object Cardinality}
\label{sec:cardinality}

\begin{figure}[t]
  \centering
  \begin{minipage}{0.48\linewidth}
    \centering
    \includegraphics[width=\linewidth]{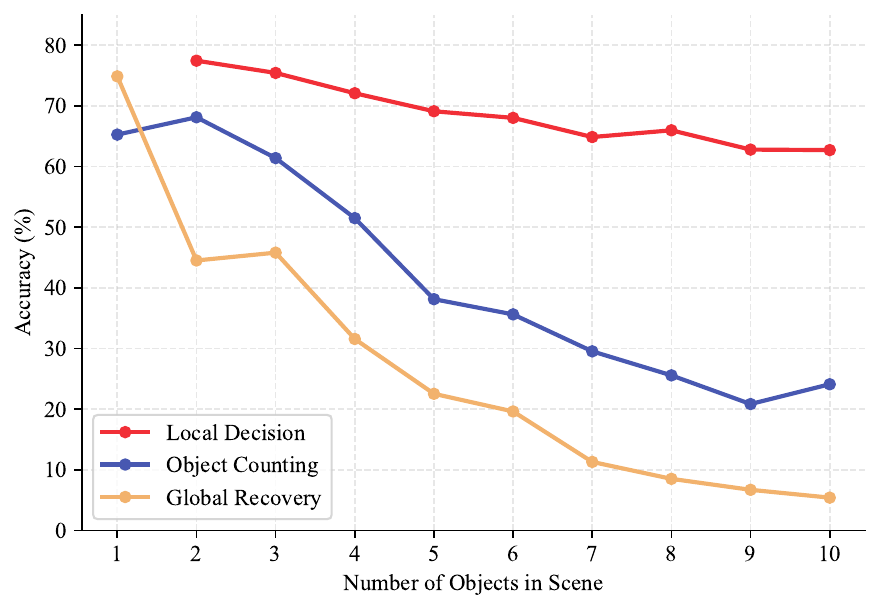}
    \caption{Accuracy vs. Object Count per Reasoning Category.}
    \label{fig:obj_count}
  \end{minipage}
  \hfill
  \begin{minipage}{0.48\linewidth}
    \centering
    \includegraphics[width=\linewidth]{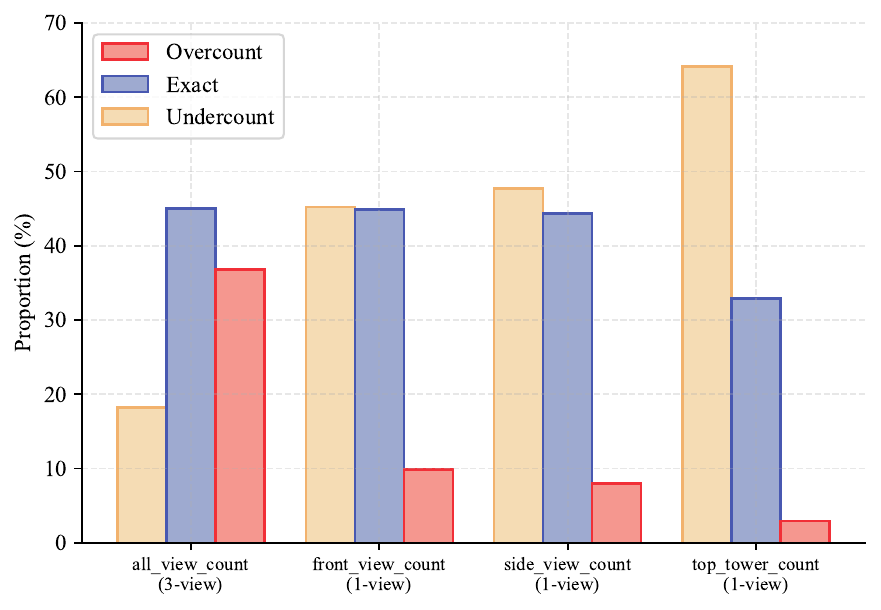}
    \caption{Counting error by sub-type.}
    \label{fig:type_counting_error}
  \end{minipage}
\end{figure}

\textit{Object count is the primary driver of structural reasoning failure,
with impact proportional to how completely a task depends on full-scene enumeration.}
\Cref{fig:obj_count} plots per-category accuracy as a function of scene object count.
The three curves diverge sharply and directly mirror the capability hierarchy established in \cref{sec:category}:
Local Decision degrades only modestly and remains above 60\% even at maximum density.
Object Counting declines moderately from 65.28\% to 24.12\%;
Global Recovery collapses from 74.88\% to 5.43\%;
The divergence follows from what each task actually requires.
Local Decision compares two specified objects;
other objects in the scene simply do not enter the judgment,
so adding more of them has little effect.
Object Counting and Global Recovery, by contrast,
must account for every object present:
each additional object is one more entity to track,
and any single miss propagates into the final answer.

\textit{Error analysis on Object Counting reveals two independent failure mechanisms whose direction reverses with the number of available viewpoints.}
Object Counting is the only category that produces numerical predictions,
enabling analysis of not just whether models fail but how they fail.
\Cref{fig:type_counting_error} shows that error direction depends critically on how many viewpoints are available.
Under single-view conditions, undercounting dominates.
Since front/side\_view\_count tasks only require counting visible objects in the given image, no spatial inference about hidden objects is needed;
the undercounting therefore reflects a failure of fine-grained visual perception rather than a reasoning limitation.
The top-view tower counting sub-task is particularly revealing.
Counting towers from above is geometrically straightforward,
yet models perform worst here (exact accuracy 32.92\%, MAE 1.83,
compared to roughly 44\% and 0.9 for front and side views).
This disparity likely stems from training distribution bias:
top-down perspectives are underrepresented in pretraining data,
leaving models poorly calibrated for this viewpoint.
The three-view condition tells a different story.
When all three perspectives are provided simultaneously,
overcounting rises to 36.78\% while undercounting drops to 18.19\%,
with MAE at 0.90.
With all viewpoints available, visibility is no longer a bottleneck;
the overcounting instead indicates that models fail to establish object identity across views,
registering the same physical object once per perspective.
The two failure modes are opposite in direction,
pointing to two independent perceptual breakdowns:
insufficient visual perception causes undercounting in single-view settings,
while failed cross-view correspondence causes overcounting when multiple perspectives are combined.

\subsubsection{Effect of Occlusion}
\label{sec:occlusion_specific}

\begin{table}[t] 
    \centering                
    \begin{minipage}[t]{0.58\linewidth} 
        \centering
        \makeatletter\def\@captype{figure}\makeatother 
        \includegraphics[width=0.95\linewidth]{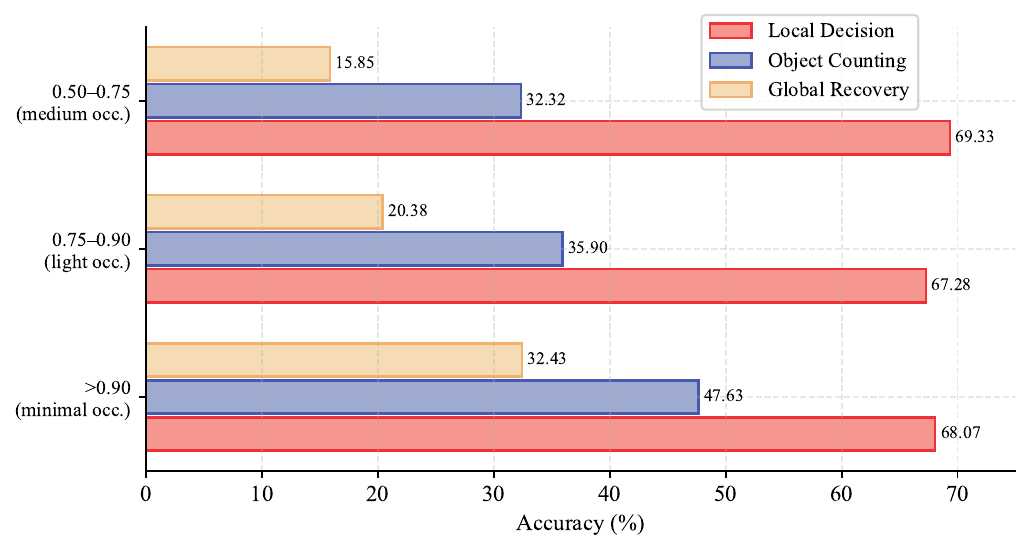}
        \caption{Accuracy by scene-level visibility ratio.}
        \label{fig:visibility}
    \end{minipage}
    \hfill
    \begin{minipage}[t]{0.40\linewidth} 
        \centering
        \raisebox{40pt}{
            \begin{minipage}{\linewidth} 
                \centering
                \caption{Average accuracy across 18 MLLMs on occlusion-specific task types.}
                \label{tab:occlusion}
                \resizebox{\linewidth}{!}{
                    \begin{tabular}{lcc}
                    \toprule 
                    Question Type                  & Level 2 & Level 4 \\
                    \midrule
                    front\_hidden\_count  & 46.56   & 28.22   \\
                    side\_hidden\_count   & 49.78   & 27.55   \\
                    front\_occluded\_ref  & 49.71   & 32.98   \\
                    side\_occluded\_ref   & 47.07   & 35.59   \\
                    \bottomrule
                    \end{tabular}
                }
            \end{minipage}
        }
    \end{minipage}
\end{table}

\textit{Occlusion selectively impairs tasks that require complete scene enumeration,
while leaving pairwise spatial judgments intact.}
\Cref{fig:visibility} stratifies accuracy by scene-level mean visibility ratio,
binned into three occlusion levels.
Local Decision accuracy is essentially flat across all bins (67.28\%--69.33\%),
confirming that relative spatial judgments between two objects are unaffected by the occlusion state of other scene objects.
Object Counting declines from 47.63\% to 32.32\% and Global Recovery shows the steepest drop, from 32.43\% to 15.85\%.
The dissociation follows directly from task structure.
Local Decision can be resolved from any visible pair of objects and is therefore robust to partial occlusion.
Object Counting and Global Recovery require access to the complete set of scene objects,
so a single occluded object is sufficient to introduce an error,
and the probability of at least one such failure grows with both object count and occlusion density.

The decrements in \cref{fig:visibility} measure only the downstream consequences of occlusion on counting and recovery performance.
To probe whether models also struggle when occlusion is queried explicitly,
Levels~2 and~4 include four question types that ask about occluded objects directly (\cref{tab:occlusion}).
Hidden\_count asks how many objects are completely invisible in a given view,
requiring the model to identify absent objects by consulting other viewpoints.
Occluded\_ref asks about the visual characteristics of an object partially occluded by a named reference, again requiring cross-view inference.
Both types yield consistently poor results (46\%--50\% at Level~2;
below 36\% at Level~4), confirming that models struggle with occlusion whether it affects task performance as a side effect or is the direct target of the question.

\subsubsection{CoT Prompting Does Not Resolve Structural Failures}

\begin{figure}[t]
  \centering
  \includegraphics[width=\linewidth]{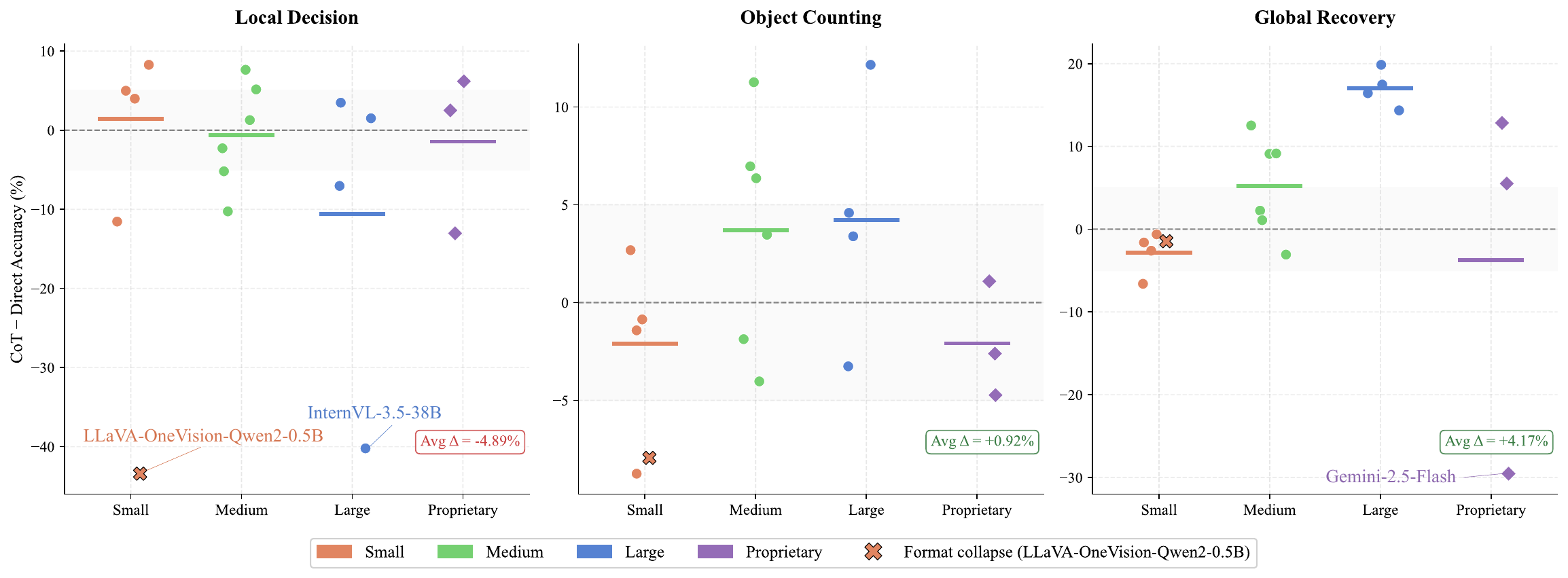}
  \caption{
    \textbf{CoT $-$ Direct accuracy gap by reasoning category and
    model scale.}
    Each point is one model; horizontal bars show group means
    (format-collapsed models excluded from means).
    For open-source models, Small denotes models with less than 3B parameters, Medium refers to those with 4B–8B parameters, and Large indicates models with 14B parameters or more.
  }
  \label{fig:cot}
\end{figure}

The analyses above identify two independent sources of structural reasoning failure:
increasing object cardinality overwhelms scene tracking capacity,
and occlusion removes objects from direct observation.
A natural question is whether providing more reasoning steps can offset these limitations.
\Cref{fig:cot} reports the CoT effect across three categories and model scales.
The overall effect is nearly zero ($\Delta = -0.16\%$),
but this aggregate masks qualitatively distinct per-category patterns.

\paragraph{Local Decision ($\Delta = -4.89\%$).}
Local Decision shows the largest average decline, but the mechanism varies across models.
LLaVA-OneVision-Qwen2-0.5B exhibits a near-total collapse under CoT (accuracy $\approx 0\%$ across all levels),
which is not a reasoning degradation but an \emph{instruction-format failure}:
the model cannot follow the extended CoT response format and produces unparseable outputs,
establishing a minimum capability threshold below which CoT cannot function at all.
Among capable models, several (notably InternVL3.5-38B , $-40.22\%$) show large drops consistent with CoT-induced overthinking:
on tasks requiring direct perceptual judgment,
extended reasoning chains cause models to reconsider and reverse initially correct answers.

\paragraph{Object Counting ($\Delta = +0.92\%$).}
Object Counting is essentially unchanged by CoT, with no systematic directional effect.
Extended reasoning steps neither compensate for missing perceptual information nor introduce new errors, suggesting that the fundamental limitation of incomplete scene inventory is orthogonal to the number of reasoning steps taken.

\paragraph{Global Recovery ($\Delta = +4.17\%$).}
Global Recovery is the only category that shows a net positive response to CoT,
and the benefit is strongly capability-gated.
All four large open-source models improve under CoT (group mean $+17.03\%$),
while all small models deteriorate (group mean $-2.86\%$),
with medium-scale models showing intermediate, mostly positive responses.
This pattern suggests that CoT benefits Global Recovery \emph{only when} the model already possesses sufficient baseline representational capacity:
when some scene structure is accessible to begin with,
explicit step-by-step reasoning can help organize it.
Gemini-2.5-Flash is a notable exception:
despite achieving the highest Direct accuracy on Global Recovery,
it degrades sharply under CoT ($-29.54\%$),
possibly related to architectural choices optimized for concise generation that may not align well with extended reasoning formats.
Even for the models that benefit most, the gains remain insufficient:
large open-source models still fall well below 30\% on Global Recovery at Level~4 even with CoT, far short of useful performance.

These three patterns converge on a single conclusion.
Local Decision is impaired or collapses under CoT because the task is simple enough that extra reasoning actively interferes or cannot be executed.
Object Counting is unaffected because reasoning steps cannot recover information absent from the model's perceptual representation.
Global Recovery benefits conditionally from CoT, but only for models that already have partial structural understanding, and even then the gains leave an enormous gap to human-level performance.
\textbf{The root limitation is the inability to form a complete and consistent cross-view spatial representation, and this is a representational failure that no prompting strategy can substitute for.}

% ----------------------------------------------------------------
\section{Limitations}
% ----------------------------------------------------------------

TriViewBench is constructed from synthetic scenes, affording precise control over complexity variables but limiting generalization to real-world imagery.
Synthetic renderings do not replicate photometric variation,
texture diversity, and environmental clutter of natural scenes,
and whether the capability hierarchy and failure modes identified here transfer to real-world multi-view settings remains open.

The benchmark adopts a fixed three-view configuration (front, side, top), which does not capture the continuous viewpoint variation encountered in practice.
Evaluating structural reasoning under arbitrary camera placement or with fewer than three views is outside the current scope.

The three reasoning categories target a specific slice of multi-view structural understanding and do not cover all relevant capabilities.
Tasks such as open-ended scene description, 3D trajectory reasoning,
or scene graph construction fall outside the current design.

Finally, the CoT evaluation uses a single prompt format for each model. It remains open whether prompt design choices, including few-shot demonstrations or formatting tailored to model families, could alter the capability-gating pattern observed in Global Recovery, and this merits further investigation.

% ----------------------------------------------------------------
\section{Conclusion}
% ----------------------------------------------------------------

We introduced TriViewBench, a controlled benchmark for diagnosing multi-view structural reasoning in MLLMs under explicitly defined complexity.
By independently varying object count and occlusion across four levels, the benchmark enables direct attribution of performance changes to specific structural factors.

Evaluation of 18 MLLMs reveals four consistent findings.
First, all 18 models exhibit an identical capability hierarchy (Local Decision $>$ Object Counting $>$ Global Recovery) without exception across architectures, scales, and families,
establishing this as a systematic property of current MLLMs rather than a model-specific artifact.
Second, performance degrades monotonically with complexity at markedly different rates: Local Decision tasks decline modestly (12.11\% relative drop), Object Counting degrades substantially (59.14\% relative drop), and Global Recovery deteriorates severely (80.02\% relative drop).
Third, error direction analysis on Object Counting uncovers two mechanistically independent failure modes: single-view occlusion blindness, in which models fail to detect objects partially hidden from a single perspective, and cross-view identity confusion, in which models count the same physical object multiple times across views.
These modes are comparable in magnitude but opposite in direction,
pointing to distinct perceptual bottlenecks rather than a single underlying deficit.
Fourth, CoT prompting yields near-zero overall benefit ($\Delta = -0.16\%$), and its effect on Global Recovery is strongly capability-gated: large open-source models ($\geq$14B) improve by an average of $+17.03\%$ while small models ($\leq$3B) uniformly deteriorate.
Even with CoT, no model approaches human-level performance,
with a gap exceeding 30 percentage points at the hardest level.

These findings collectively indicate that the root limitation is not a lack of reasoning steps but the inability to form complete and consistent cross-view spatial representations.
TriViewBench provides a controlled diagnostic framework for guiding future progress on structural multi-view understanding.

% \section*{Acknowledgements}
% Please insert your acknowledgments here.

\bibliographystyle{splncs04}
\bibliography{main}
\newpage
\appendix
\section{Per-Model Detailed Results}
\label{sec:appendix_results}

Tables~\ref{tab:appendix_direct} and~\ref{tab:appendix_cot} report
per-model accuracy broken down by reasoning category and complexity
level under Direct and CoT prompting respectively.
LD = Local Decision, OC = Object Counting, GR = Global Recovery.
L1--L4 denote increasing complexity levels
(L1: LowCount\_LowOcclusion; L2: LowCount\_HighOcclusion;
L3: HighCount\_LowOcclusion; L4: HighCount\_HighOcclusion).
Avg is computed as a micro-average across all four levels.
$\Delta$ in Table~\ref{tab:appendix_cot} denotes the difference
in Avg between CoT and Direct prompting; positive values
(shown in \textbf{bold}) indicate improvement under CoT.

% ----------------------------------------------------------------
{
\setlength{\tabcolsep}{4pt}
\setlength\LTleft{\fill}
\setlength\LTright{\fill}
\begin{longtable}{llrrrrr}
\caption{Per-model accuracy (\%) under Direct prompting.}
\label{tab:appendix_direct} \\
\toprule
\textbf{Model} & \textbf{Cat.} & \textbf{L1} & \textbf{L2} &
\textbf{L3} & \textbf{L4} & \textbf{Avg} \\
\midrule
\endfirsthead
\multicolumn{7}{l}{\small\textit{(continued from previous page)}} \\
\toprule
\textbf{Model} & \textbf{Cat.} & \textbf{L1} & \textbf{L2} &
\textbf{L3} & \textbf{L4} & \textbf{Avg} \\
\midrule
\endhead
\midrule
\multicolumn{7}{r}{\small\textit{(continued on next page)}} \\
\endfoot
\bottomrule
\endlastfoot

\rowcolor[gray]{.95} \multicolumn{7}{l}{\textit{Proprietary}} \\
\multirow{3}{*}{GPT-4o} & LD & 81.73 & 77.06 & 72.25 & 71.30 & 74.52 \\*
 & OC & 87.20 & 66.74 & 45.30 & 35.21 & 58.52 \\*
 & GR & 88.20 & 68.10 & 26.03 & 9.05 & 47.48 \\
\midrule
\multirow{3}{*}{Claude-3.7-Sonnet} & LD & 86.25 & 81.80 & 80.38 & 78.43 & 80.99 \\*
 & OC & 83.05 & 72.40 & 67.36 & 50.45 & 68.23 \\*
 & GR & 86.40 & 64.71 & 43.80 & 31.19 & 56.42 \\
\midrule
\multirow{3}{*}{Gemini-2.5-Flash} & LD & 94.54 & 89.20 & 90.24 & 87.09 & 89.65 \\*
 & OC & 90.60 & 82.86 & 79.80 & 61.52 & 78.59 \\*
 & GR & 92.00 & 83.94 & 72.31 & 54.53 & 75.51 \\
\midrule
\rowcolor[gray]{.95} \multicolumn{7}{l}{\textit{Open-source}} \\
\multirow{3}{*}{LLaVA-OneVision-Qwen2-0.5B} & LD & 45.01 & 41.63 & 46.99 & 40.61 & 43.42 \\*
 & OC & 14.00 & 19.00 & 0.93 & 7.04 & 10.06 \\*
 & GR & 6.20 & 0.90 & 0.00 & 0.00 & 1.82 \\
\midrule
\multirow{3}{*}{LLaVA-OneVision-Qwen2-7B} & LD & 56.12 & 52.67 & 43.35 & 42.60 & 47.23 \\*
 & OC & 47.20 & 20.76 & 8.94 & 8.60 & 21.52 \\*
 & GR & 28.00 & 6.79 & 0.00 & 2.21 & 9.41 \\
\midrule
\multirow{3}{*}{Qwen2.5-VL-3B-Instruct} & LD & 60.08 & 52.79 & 38.95 & 46.57 & 47.81 \\*
 & OC & 47.95 & 19.46 & 4.65 & 9.41 & 20.54 \\*
 & GR & 30.60 & 8.37 & 0.00 & 1.61 & 10.30 \\
\midrule
\multirow{3}{*}{Qwen2.5-VL-7B-Instruct} & LD & 62.71 & 66.26 & 44.40 & 50.09 & 54.10 \\*
 & OC & 70.00 & 39.03 & 16.74 & 14.24 & 35.06 \\*
 & GR & 42.20 & 1.13 & 0.00 & 0.80 & 11.44 \\
\midrule
\multirow{3}{*}{Qwen2.5-VL-32B-Instruct} & LD & 78.15 & 65.90 & 54.07 & 55.14 & 60.83 \\*
 & OC & 68.55 & 47.74 & 32.70 & 26.21 & 43.80 \\*
 & GR & 50.20 & 19.23 & 5.17 & 1.61 & 19.19 \\
\midrule
\multirow{3}{*}{Qwen3-VL-2B-Instruct} & LD & 72.88 & 75.73 & 70.81 & 68.32 & 71.49 \\*
 & OC & 65.60 & 44.29 & 30.89 & 24.75 & 41.41 \\*
 & GR & 26.00 & 11.99 & 0.00 & 1.81 & 9.98 \\
\midrule
\multirow{3}{*}{Qwen3-VL-4B-Instruct} & LD & 72.13 & 75.97 & 69.19 & 67.42 & 70.67 \\*
 & OC & 68.65 & 54.64 & 34.81 & 31.09 & 47.20 \\*
 & GR & 70.40 & 47.29 & 23.55 & 14.49 & 38.85 \\
\midrule
\multirow{3}{*}{Qwen3-VL-8B-Instruct} & LD & 86.82 & 87.26 & 83.83 & 81.95 & 84.49 \\*
 & OC & 77.60 & 60.69 & 48.81 & 38.98 & 56.49 \\*
 & GR & 79.60 & 44.80 & 21.69 & 11.47 & 39.42 \\
\midrule
\multirow{3}{*}{Qwen3-VL-32B-Instruct} & LD & 93.03 & 87.01 & 87.37 & 83.21 & 86.83 \\*
 & OC & 85.65 & 66.23 & 46.80 & 29.68 & 56.94 \\*
 & GR & 80.20 & 56.33 & 12.60 & 14.69 & 40.77 \\
\midrule
\multirow{3}{*}{InternVL3.5-1B} & LD & 58.57 & 57.28 & 51.20 & 52.89 & 54.28 \\*
 & OC & 50.10 & 31.84 & 16.94 & 18.41 & 29.37 \\*
 & GR & 23.40 & 4.07 & 0.21 & 0.20 & 7.12 \\
\midrule
\multirow{3}{*}{InternVL3.5-2B} & LD & 53.30 & 54.61 & 57.22 & 50.45 & 53.88 \\*
 & OC & 60.60 & 34.33 & 22.00 & 18.71 & 34.02 \\*
 & GR & 45.20 & 17.87 & 0.00 & 0.00 & 15.86 \\
\midrule
\multirow{3}{*}{InternVL3.5-4B} & LD & 80.04 & 79.49 & 76.75 & 77.71 & 78.19 \\*
 & OC & 48.40 & 37.22 & 27.58 & 16.60 & 32.37 \\*
 & GR & 20.80 & 15.38 & 12.81 & 6.24 & 13.78 \\
\midrule
\multirow{3}{*}{InternVL3.5-8B} & LD & 76.46 & 74.88 & 73.21 & 69.77 & 73.00 \\*
 & OC & 37.85 & 30.77 & 24.95 & 12.93 & 26.53 \\*
 & GR & 15.00 & 21.72 & 10.74 & 12.47 & 14.82 \\
\midrule
\multirow{3}{*}{InternVL3.5-14B} & LD & 80.41 & 77.67 & 72.25 & 69.95 & 74.03 \\*
 & OC & 51.30 & 40.72 & 30.06 & 21.38 & 35.79 \\*
 & GR & 23.80 & 16.06 & 6.82 & 5.84 & 13.10 \\
\midrule
\multirow{3}{*}{InternVL3.5-38B} & LD & 85.88 & 77.06 & 73.88 & 70.22 & 75.29 \\*
 & OC & 45.35 & 37.50 & 29.34 & 23.99 & 34.00 \\*
 & GR & 70.20 & 22.40 & 4.34 & 7.24 & 26.37 \\
\end{longtable}
}
\clearpage
% ----------------------------------------------------------------
{
\setlength{\tabcolsep}{4pt}
\setlength\LTleft{\fill}
\setlength\LTright{\fill}
\begin{longtable}{llrrrrrr}
\caption{Per-model accuracy (\%) under CoT prompting and accuracy
gap $\Delta$ (CoT $-$ Direct).}
\label{tab:appendix_cot} \\
\toprule
\textbf{Model} & \textbf{Cat.} & \textbf{L1} & \textbf{L2} &
\textbf{L3} & \textbf{L4} & \textbf{Avg} & $\boldsymbol{\Delta}$ \\
\midrule
\endfirsthead
\multicolumn{8}{l}{\small\textit{(continued from previous page)}} \\
\toprule
\textbf{Model} & \textbf{Cat.} & \textbf{L1} & \textbf{L2} &
\textbf{L3} & \textbf{L4} & \textbf{Avg} & $\boldsymbol{\Delta}$ \\
\midrule
\endhead
\midrule
\multicolumn{8}{r}{\small\textit{(continued on next page)}} \\
\endfoot
\bottomrule
\endlastfoot

\rowcolor[gray]{.95} \multicolumn{8}{l}{\textit{Proprietary}} \\
\multirow{3}{*}{GPT-4o} & LD & 82.49 & 82.40 & 80.48 & 78.79 & 80.70 & \textbf{+6.18} \\*
 & OC & 89.65 & 73.30 & 47.31 & 29.18 & 59.61 & \textbf{+1.09} \\*
 & GR & 90.00 & 71.49 & 36.36 & 15.49 & 52.99 & \textbf{+5.51} \\
\midrule
\multirow{3}{*}{Claude-3.7-Sonnet} & LD & 89.08 & 86.04 & 82.78 & 79.60 & 83.49 & \textbf{+2.50} \\*
 & OC & 75.25 & 70.48 & 67.77 & 49.50 & 65.61 & -2.62 \\*
 & GR & 91.80 & 78.05 & 66.32 & 41.65 & 69.27 & \textbf{+12.85} \\
\midrule
\multirow{3}{*}{Gemini-2.5-Flash} & LD & 76.46 & 72.33 & 81.24 & 75.54 & 76.62 & -13.03 \\*
 & OC & 79.90 & 74.10 & 79.44 & 62.12 & 73.86 & -4.73 \\*
 & GR & 47.40 & 53.85 & 48.97 & 34.61 & 45.97 & -29.54 \\
\midrule
\rowcolor[gray]{.95} \multicolumn{8}{l}{\textit{Open-source}} \\
\multirow{3}{*}{LLaVA-OneVision-Qwen2-0.5B} & LD & 0.00 & 0.00 & 0.00 & 0.00 & 0.00 & -43.42 \\*
 & OC & 4.30 & 4.36 & 0.00 & 0.00 & 2.12 & -7.94 \\*
 & GR & 0.60 & 0.00 & 0.21 & 0.60 & 0.36 & -1.46 \\
\midrule
\multirow{3}{*}{LLaVA-OneVision-Qwen2-7B} & LD & 59.13 & 54.13 & 51.39 & 48.83 & 52.39 & \textbf{+5.16} \\*
 & OC & 51.70 & 28.90 & 9.87 & 9.36 & 24.99 & \textbf{+3.47} \\*
 & GR & 13.40 & 9.73 & 0.62 & 1.81 & 6.34 & -3.07 \\
\midrule
\multirow{3}{*}{Qwen2.5-VL-3B-Instruct} & LD & 59.32 & 54.85 & 51.00 & 46.66 & 51.80 & \textbf{+3.99} \\*
 & OC & 51.70 & 26.92 & 6.40 & 7.65 & 23.22 & \textbf{+2.68} \\*
 & GR & 33.00 & 3.17 & 0.00 & 1.41 & 9.67 & -0.63 \\
\midrule
\multirow{3}{*}{Qwen2.5-VL-7B-Instruct} & LD & 64.03 & 66.99 & 59.14 & 59.21 & 61.74 & \textbf{+7.64} \\*
 & OC & 60.60 & 40.27 & 21.02 & 11.17 & 33.19 & -1.87 \\*
 & GR & 38.60 & 11.09 & 2.89 & 1.41 & 13.68 & \textbf{+2.24} \\
\midrule
\multirow{3}{*}{Qwen2.5-VL-32B-Instruct} & LD & 65.54 & 69.42 & 63.64 & 60.56 & 64.31 & \textbf{+3.48} \\*
 & OC & 74.15 & 51.64 & 38.95 & 24.14 & 47.19 & \textbf{+3.39} \\*
 & GR & 73.00 & 39.82 & 22.11 & 7.44 & 35.62 & \textbf{+16.43} \\
\midrule
\multirow{3}{*}{Qwen3-VL-2B-Instruct} & LD & 70.62 & 64.56 & 57.13 & 54.06 & 59.95 & -11.54 \\*
 & OC & 55.70 & 32.47 & 26.24 & 15.90 & 32.66 & -8.75 \\*
 & GR & 28.20 & 2.94 & 0.00 & 1.41 & 8.37 & -1.61 \\
\midrule
\multirow{3}{*}{Qwen3-VL-4B-Instruct} & LD & 71.37 & 75.24 & 74.55 & 67.33 & 71.95 & \textbf{+1.28} \\*
 & OC & 78.45 & 60.07 & 44.16 & 31.89 & 53.56 & \textbf{+6.36} \\*
 & GR & 83.00 & 62.67 & 29.55 & 17.71 & 48.00 & \textbf{+9.15} \\
\midrule
\multirow{3}{*}{Qwen3-VL-8B-Instruct} & LD & 87.38 & 83.74 & 83.06 & 77.80 & 82.21 & -2.28 \\*
 & OC & 73.25 & 56.56 & 45.30 & 34.86 & 52.46 & -4.03 \\*
 & GR & 74.20 & 45.02 & 25.41 & 17.30 & 40.51 & \textbf{+1.09} \\
\midrule
\multirow{3}{*}{Qwen3-VL-32B-Instruct} & LD & 83.80 & 77.79 & 83.54 & 75.81 & 79.79 & -7.04 \\*
 & OC & 85.15 & 65.61 & 55.79 & 39.74 & 61.53 & \textbf{+4.59} \\*
 & GR & 87.40 & 62.90 & 47.11 & 23.54 & 55.12 & \textbf{+14.35} \\
\midrule
\multirow{3}{*}{InternVL3.5-1B} & LD & 61.39 & 59.47 & 58.85 & 58.48 & 59.26 & \textbf{+4.98} \\*
 & OC & 49.80 & 28.45 & 20.30 & 15.14 & 28.51 & -0.86 \\*
 & GR & 14.00 & 3.62 & 0.00 & 0.20 & 4.52 & -2.60 \\
\midrule
\multirow{3}{*}{InternVL3.5-2B} & LD & 66.29 & 66.38 & 60.10 & 58.94 & 62.14 & \textbf{+8.26} \\*
 & OC & 56.60 & 31.62 & 23.55 & 18.16 & 32.61 & -1.41 \\*
 & GR & 27.60 & 7.47 & 0.00 & 1.41 & 9.26 & -6.60 \\
\midrule
\multirow{3}{*}{InternVL3.5-4B} & LD & 70.24 & 69.54 & 67.46 & 66.06 & 67.93 & -10.26 \\*
 & OC & 58.65 & 46.04 & 31.82 & 21.28 & 39.34 & \textbf{+6.97} \\*
 & GR & 47.60 & 26.70 & 10.12 & 7.04 & 22.88 & \textbf{+9.10} \\
\midrule
\multirow{3}{*}{InternVL3.5-8B} & LD & 73.45 & 72.82 & 67.56 & 61.64 & 67.82 & -5.18 \\*
 & OC & 62.20 & 42.93 & 27.84 & 18.41 & 37.81 & \textbf{+11.28} \\*
 & GR & 52.60 & 28.51 & 17.98 & 10.06 & 27.35 & \textbf{+12.53} \\
\midrule
\multirow{3}{*}{InternVL3.5-14B} & LD & 77.97 & 79.00 & 75.31 & 72.02 & 75.54 & \textbf{+1.51} \\*
 & OC & 50.15 & 37.56 & 26.91 & 15.79 & 32.53 & -3.26 \\*
 & GR & 65.40 & 37.10 & 19.63 & 9.66 & 32.97 & \textbf{+19.87} \\
\midrule
\multirow{3}{*}{InternVL3.5-38B} & LD & 34.84 & 33.74 & 36.75 & 34.57 & 35.06 & -40.23 \\*
 & OC & 66.60 & 52.32 & 37.91 & 28.12 & 46.15 & \textbf{+12.15} \\*
 & GR & 75.20 & 54.30 & 29.34 & 17.10 & 43.84 & \textbf{+17.47} \\

\end{longtable}
}
\clearpage

\section{Prompt Templates}
\label{sec:appendix_prompts}
Each prompt is assembled from five components: a fixed scene context,
a view description, a question-type-specific guidance string, the
question itself, and an output constraint determined by the evaluation mode. For most question types, all three views (front, side, top) are provided in order. The exceptions are front\_view\_count,
side\_view\_count, and top\_tower\_count, each of which receives only its corresponding single view, since the task is
defined within that perspective alone. We used two evaluation modes:
\textbf{Direct} and \textbf{CoT}.

% ------------------------------------------------------------------
\subsection{Evaluation Mode Templates}
% ------------------------------------------------------------------

\begin{tcolorbox}[promptbox, title=Direct Prompt Template]
\textbf{[Scene Context]}\\
{
Context: You are analyzing a 3D scene with geometric objects (cubes,
spheres, cylinders) placed on a ground plane and stacked into towers.
Objects come in two sizes (large, small) and various colors (blue,
green, red, yellow, purple, brown, gray, cyan).}

\medskip
\textbf{[View Description]} \textit{(two cases)}
\begin{itemize}
    \item Sigle-view tasks: You are provided with ONLY ONE image: the \{view_name\} view of
the scene.
    \item Multi-view tasks: You are provided with 3 images in order: 1.~Front View, 2.~Side View, 3.~Top View. You must integrate information from all views to reason about the 3D layout.
\end{itemize}

\medskip
\textbf{[Task-Specific Guidance]}\\
(see \cref{sec:task_guidance})

\medskip
\textbf{[Question]}\\
{Question: \{question\}}

\medskip
\textbf{[Output Constraint]}\\
{Constraint: Directly output the answer inside \texttt{<answer>}
tags. No reasoning.\\
Example: \texttt{<answer>3</answer>} or \texttt{<answer>behind</answer>}.}
\end{tcolorbox}

\bigskip

\begin{tcolorbox}[promptbox, title=CoT Prompt Template]
\textbf{[Scene Context, View Description, Task-Specific Guidance,
Question]}\\
\textit{(identical to Direct)}

\medskip
\textbf{[Output Constraint]}\\
{
Constraint: Please think step-by-step to analyze the spatial layout
and occlusions. Verify your reasoning across all provided views before
concluding. Structure your response as:\\
Thinking: [Analyze the objects and their relative positions...]\\
Final Answer: \texttt{<answer>your\_final\_answer</answer>} \\
(e.g., \texttt{<answer>3</answer>} or \texttt{<answer>behind</answer>})}
\end{tcolorbox}

% ------------------------------------------------------------------
\subsection{Task-Specific Guidance Strings}
\label{sec:task_guidance}
% ------------------------------------------------------------------

The \textbf{[Task-Specific Guidance]} component is selected by
question type as follows.

\bigskip

\begin{tcolorbox}[promptbox, title={all\_view\_count}]
{Task: Identify and count all distinct physical objects present in the 3D scene by integrating information from all provided views.}
\end{tcolorbox}

\begin{tcolorbox}[promptbox,
title={front\_view\_count} / {side\_view\_count}]
{Task: Count only the objects VISIBLE in this \{front / side\} view.
An object is `visible' if it has pixels > 0 shown in this specific
image. Do not count objects that are completely occluded from this
perspective.}
\end{tcolorbox}

\begin{tcolorbox}[promptbox,
title={front\_hidden\_count} / {side\_hidden\_count}]
{Task: Count objects that exist in the scene but are
COMPLETELY HIDDEN (0 pixels visible) in the \{front / side\} view. Use other views to confirm their existence.}
\end{tcolorbox}

\begin{tcolorbox}[promptbox,
title={front\_occluded\_ref} / {side\_occluded\_ref}]
{Task: Identify the target object partially or mostly hidden
behind another `blocker' object. Based on the question, output ONLY the specific attribute requested (e.g., its color or shape).\\
Possible colors: [blue, green, red, yellow, purple, brown, gray,
cyan].\\
Possible shapes: [cube, sphere, cylinder].}
\end{tcolorbox}

\begin{tcolorbox}[promptbox,
title={x/y/z_relation}]
{Task: Compare the relative positions of two unique objects
in 3D space.\\
Spatial Definition:\\
- Left/Right (X-axis): Relative to the front camera view. `Left'
means a larger X value, `Right' means a smaller X value.\\
- In front of/Behind (Y-axis): `In front of' means closer to the
front camera (larger Y value), `Behind' means further away (smaller Y
value).\\
- Above/Below (Z-axis): `Above' means a higher vertical position
(larger Z value), `Below' means lower (smaller Z value).\\
(x_relation) Choice: `left' or `right'\\
(y_relation) Choice: `in front of' or `behind'\\
(z_relation) Choice: `above' or `below'}
\end{tcolorbox}

\begin{tcolorbox}[promptbox, title={tower\_height\_max}]
{Task: Identify all vertical stacks (towers). Find the stack
containing the GREATEST NUMBER of objects and state that maximum
count. A stack is defined as objects placed directly on top of one
another.}
\end{tcolorbox}

\begin{tcolorbox}[promptbox, title={tower\_height\_list}]
{Task: Identify all vertical stacks (towers) in the scene.
Count how many objects are in each individual tower. Output the counts
as a list of integers. An object on the ground that has nothing on top
of it is considered a tower of height 1.}
\end{tcolorbox}

\begin{tcolorbox}[promptbox, title={top\_tower\_count}]
{Task: Count the total number of vertical stacks (towers) in
the scene. Each stack is counted as one `tower' regardless of its
height. An object on the ground that has nothing on top of it is
considered a tower of height 1.}
\end{tcolorbox}

\section{Answer Extraction and Evaluation Protocol}
\label{sec:appendix_eval}

\subsection{Answer Extraction}

Model outputs are parsed using a priority-based extraction pipeline.
The final extracted string is lowercased and stripped of trailing
punctuation before evaluation.

\begin{tcolorbox}[promptbox, title=Answer Extraction Priority]
\textbf{1.} Search for all \texttt{<answer>...</answer>} tag pairs (case-insensitive). If found, take the \emph{last} match.
Taking the last match prevents false positives from example tags that some models reproduce verbatim inside their reasoning trace.

\medskip
\textbf{2.} If no \texttt{<answer>} tag is found, search for
\texttt{{\textbackslash}boxed\{...\}} notation.
Content is extracted using bracket-depth counting to handle nested
braces correctly, supporting formats such as
\texttt{{\textbackslash}boxed\{right\}},
\texttt{{\textbackslash}boxed\{[2,\,2,\,1]\}}, and
\texttt{{\textbackslash}boxed\{{\textbackslash}text\{below\}\}}.
JSON double-escaped variants
(\texttt{{\textbackslash\textbackslash}boxed}) are normalised before
parsing. \texttt{{\textbackslash}text\{...\}} wrappers are stripped
from the result.
This fallback handles Gemini outputs under CoT prompting, which may format answers in \LaTeX\ boxed notation rather than
\texttt{<answer>} tags.

\medskip
\textbf{3.} If neither tag nor boxed notation is found, search for
the keyword \texttt{Final Answer:} and take all text following it.

\medskip
\textbf{4.} If none of the above match, return the full output
after stripping leading and trailing whitespace.
\end{tcolorbox}

\subsection{Answer Judgment}

Extracted predictions are compared against ground-truth answers
using one of two rules depending on question type.

\begin{tcolorbox}[promptbox, title=Judgment Rules]
\textbf{Exact match} (all question types except
\texttt{tower\_height\_list}):\\
The prediction is correct if and only if it equals the ground-truth
string after lowercasing and punctuation stripping.

\medskip
\textbf{Order-invariant list match} (\texttt{tower\_height\_list}
only):\\
All integers are extracted from both the predicted and ground-truth
strings. The prediction is correct if and only if both lists contain
the same integers in the same \emph{descending} sorted order and have
the same length.
For example, \texttt{[1, 3, 2]} and \texttt{[3, 2, 1]} are both
normalised to \texttt{[3, 2, 1]} and would be judged correct against
each other.
\end{tcolorbox}

\section{Dataset Statistics}
\label{sec:appendix_stats}

Table~\ref{tab:dataset_stats} reports the number of scenes and
questions per type at each complexity level.
Occlusion-specific types (\texttt{hidden\_count} and
\texttt{occluded\_ref}) appear only in high-occlusion levels
(L2 and L4) by design.

\begin{table}[htbp]
\centering
\caption{Number of scenes and questions per type at each
complexity level. ``--'' indicates types absent at that level.}
\label{tab:dataset_stats}
\setlength{\tabcolsep}{5pt}
\resizebox{\linewidth}{!}{
\begin{tabular}{llrrrrrr}
\toprule
\textbf{Category} & \textbf{Question Type} & \textbf{L1} & \textbf{L2} & \textbf{L3} & \textbf{L4} & \textbf{Total} \\
\midrule
\rowcolor[gray]{.92}
\multicolumn{2}{l}{\textit{Scenes}} & 500 & 442 & 484 & 497 & 1,923 \\
\rowcolor[gray]{.92}
\multicolumn{2}{l}{\textit{Questions}} & 3,031 & 3,447 & 3,465 & 4,231 & 14,174 \\
\midrule
\multirow{4}{*}{\textbf{LD}}
& tower\_height\_max     & 117 & 255 & 360 & 408 & 1,140 \\
& x\_relation   & 105 & 189 & 213 & 250 &   757 \\
& y\_relation   & 105 & 188 & 251 & 216 &   760 \\
& z\_relation   & 204 & 192 & 221 & 234 &   851 \\
\midrule
\multirow{4}{*}{\textbf{OC}}
& all\_view\_count       & 500 & 442 & 484 & 497 & 1,923 \\
& front\_view\_count     & 500 & 442 & 484 & 497 & 1,923 \\
& side\_view\_count      & 500 & 442 & 484 & 497 & 1,923 \\
& top\_tower\_count      & 500 & 442 & 484 & 497 & 1,923 \\
\midrule
\multirow{5}{*}{\textbf{GR}}
& tower\_height\_list    & 500 & 442 & 484 & 497 & 1,923 \\
& front\_hidden\_count   & --  &  76 &  -- & 125 &   201 \\
& side\_hidden\_count    & --  &  76 &  -- & 147 &   223 \\
& front\_occluded\_ref   & --  & 134 &  -- & 174 &   308 \\
& side\_occluded\_ref    & --  & 127 &  -- & 192 &   319 \\
\bottomrule
\end{tabular}}
\end{table}
\section{Example Images and Questions}
\label{sec:appendix_qual}

The remaining pages present one representative example for each of the 13 question types in TriViewBench. Each question is annotated with its question type, question and answer.

\subsection*{Local Decision}
\noindent{\includegraphics[width=\linewidth, trim=30 100 30 80, clip]{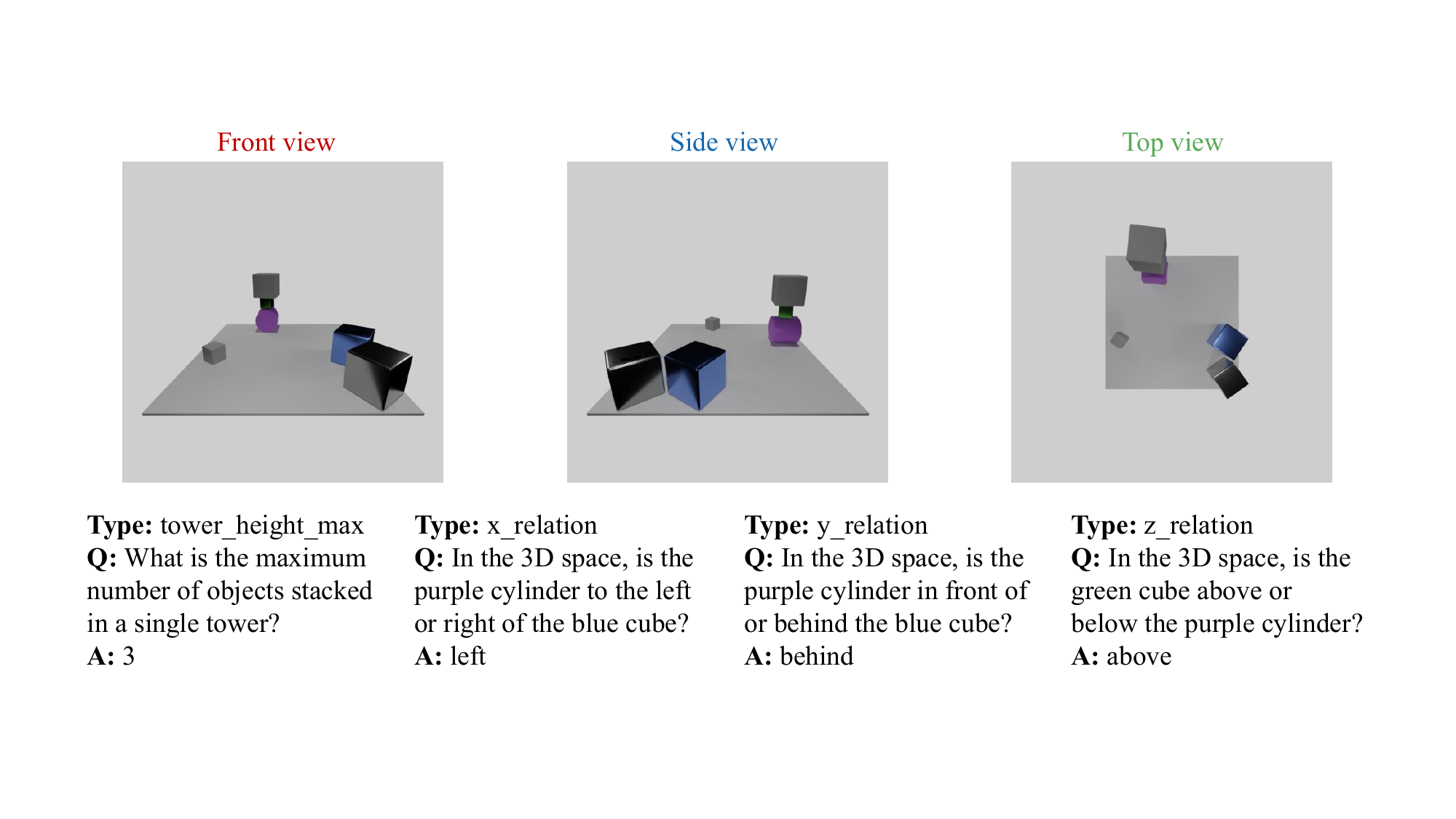}}

\subsection*{Object Counting}
\noindent{\includegraphics[width=\linewidth, trim=30 50 30 50, clip]{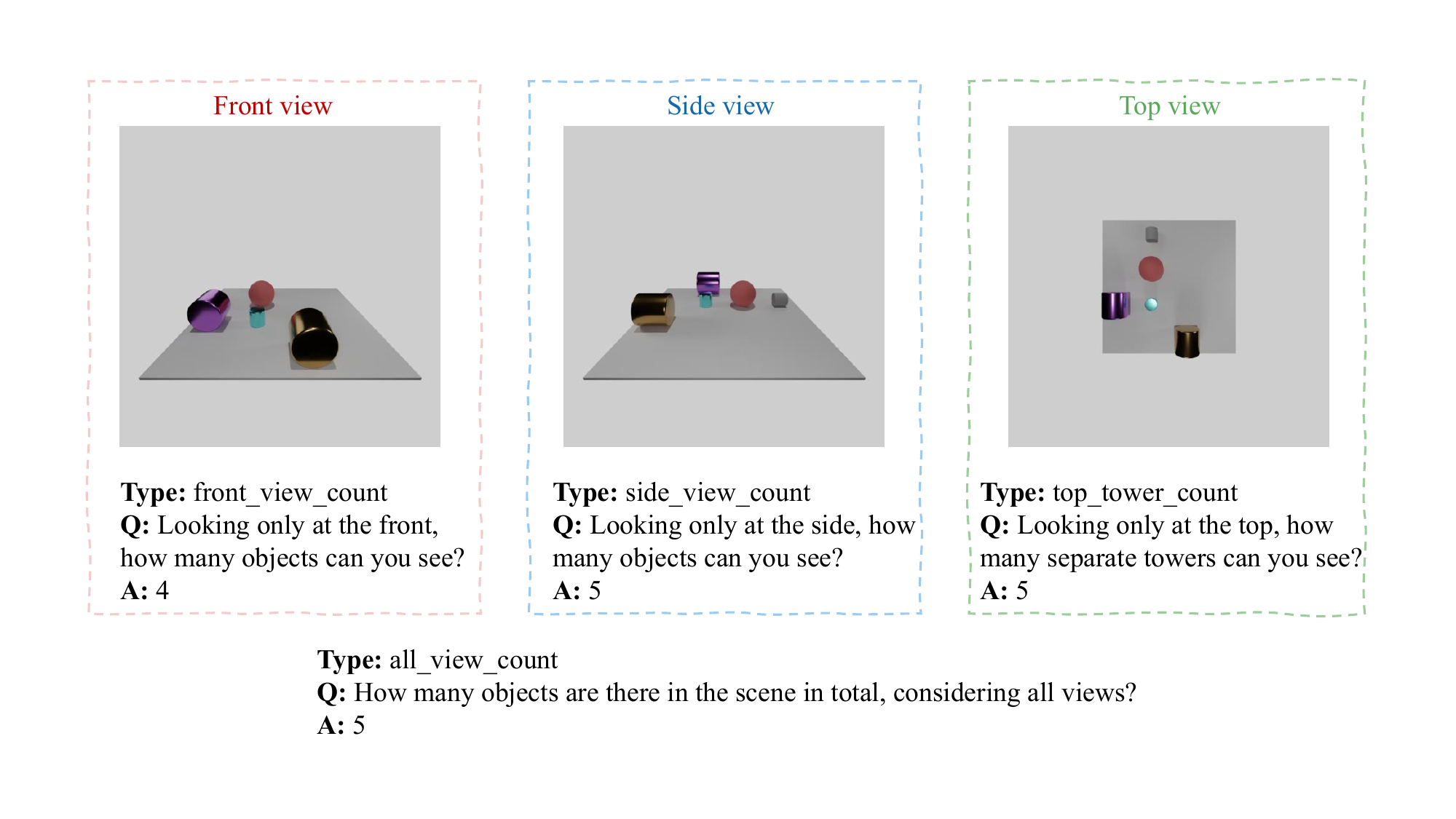}}

\subsection*{Global Recovery}
\noindent{\includegraphics[width=\linewidth, trim=30 190 30 30, clip]{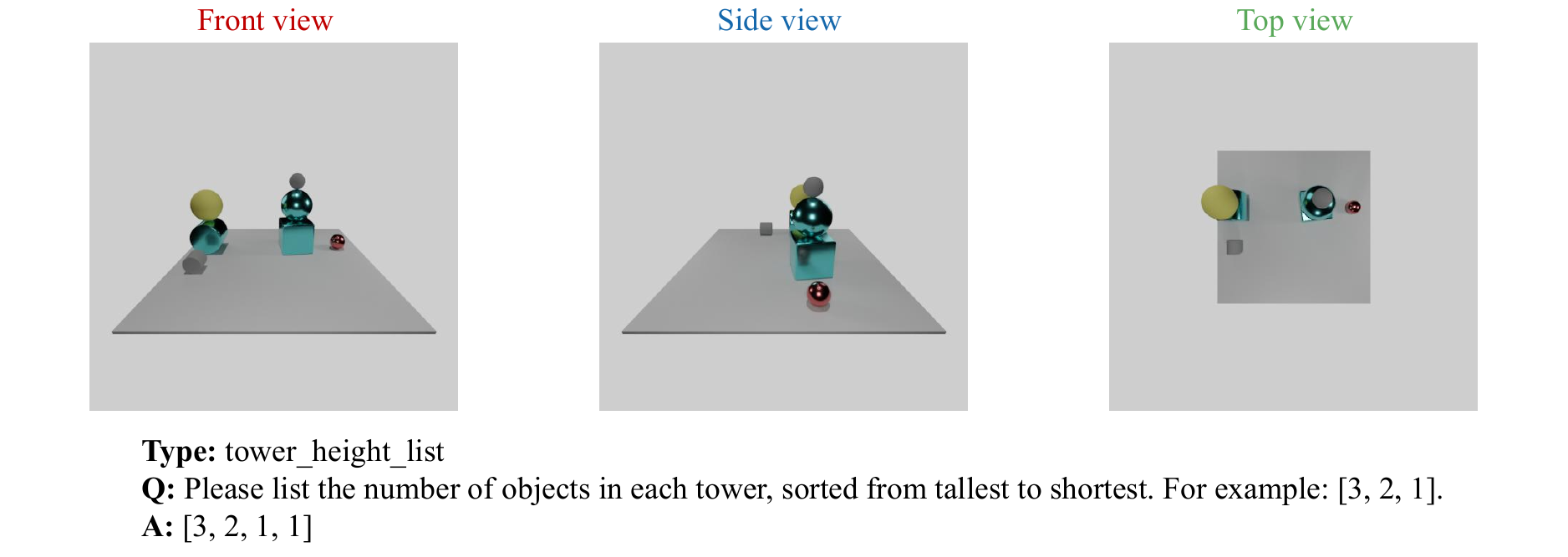}}

\begin{itemize}
    \item Occlusion-Specific Types
\end{itemize}
\noindent\includegraphics[width=\linewidth, trim=30 190 30 5, clip]{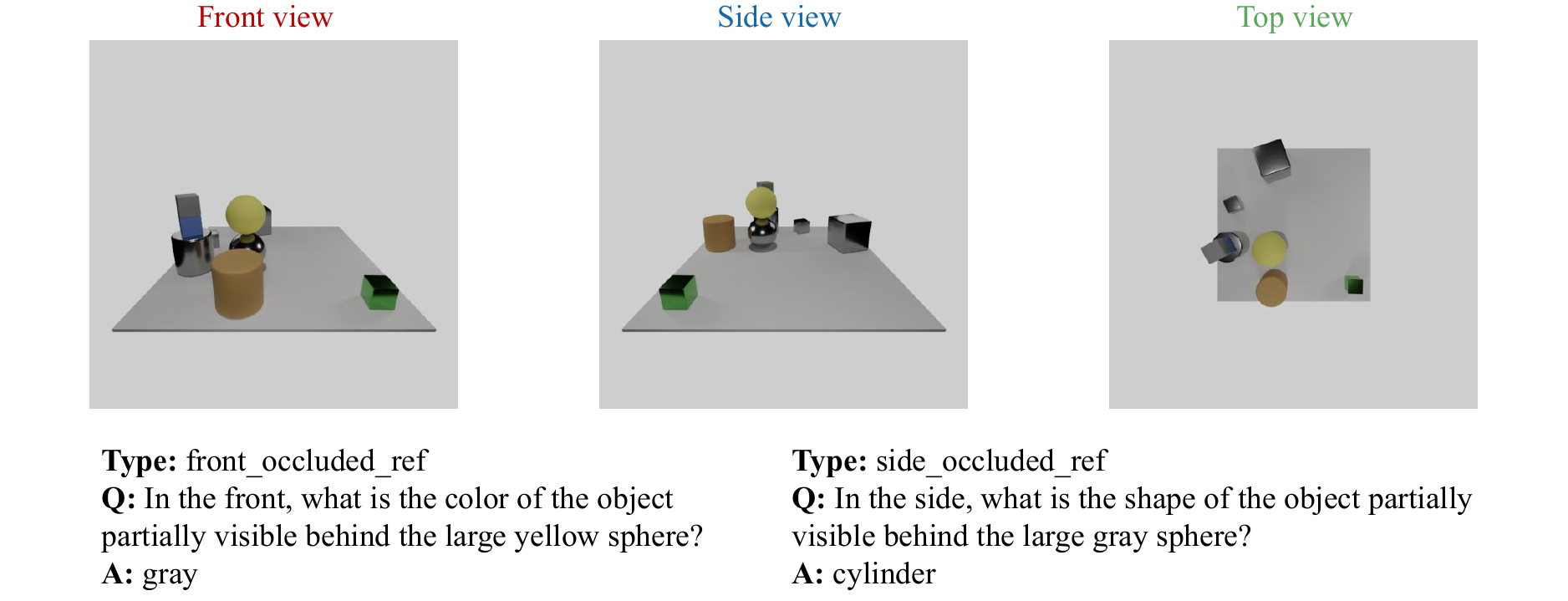}

\noindent\includegraphics[width=\linewidth, trim=30 100 30 80, clip]{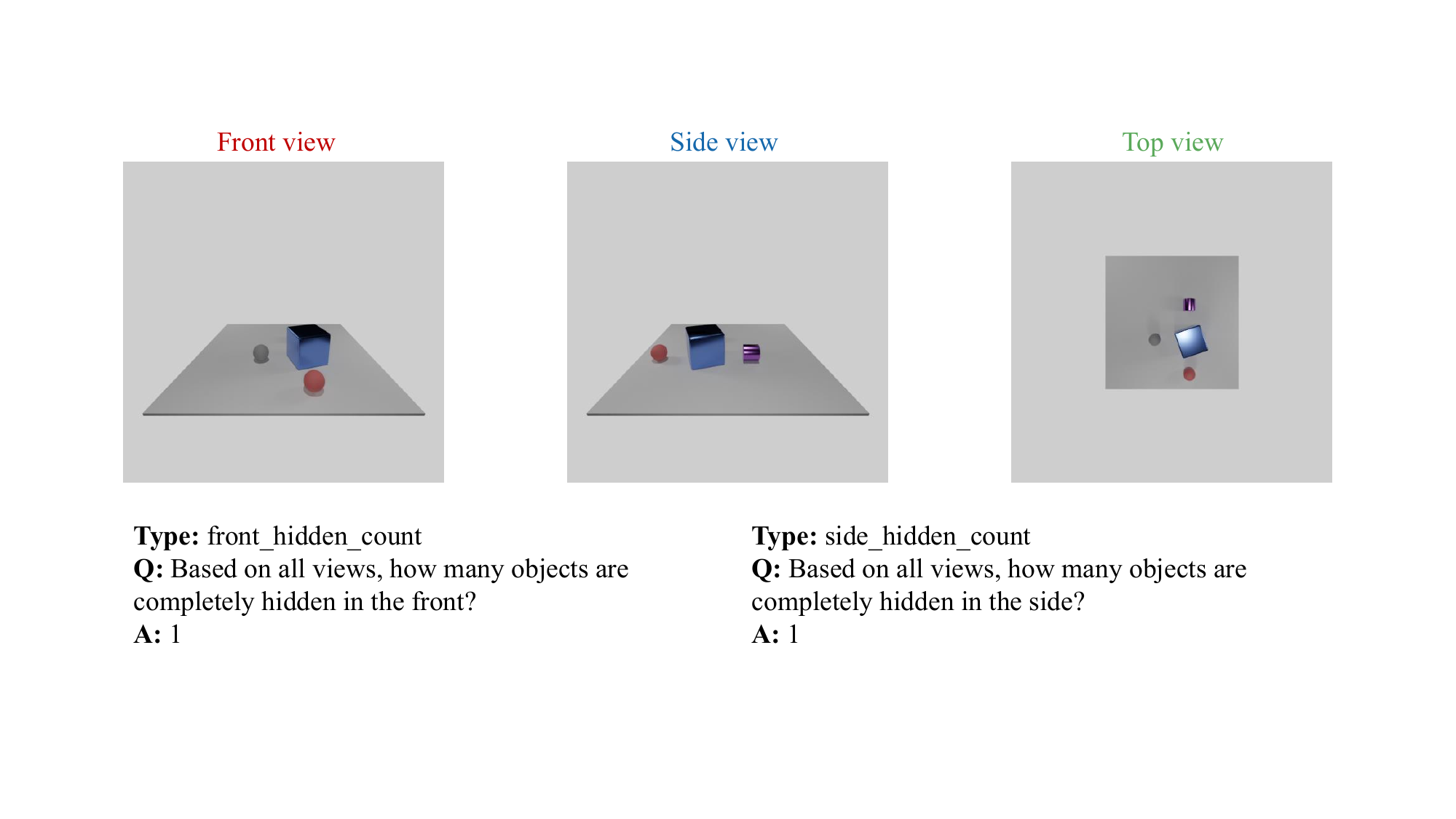}

\clearpage

\end{document}